# A Causality- and Frequency-Aware Deep Learning Framework for Wave Elevation Prediction Behind Floating Breakwaters

Zhang Jianxin[1], Jiang Lianzi[1], Han Xinyu[2,3,*], Wang Xiangrong[1]

[1] Shandong University of Science and Technology, College of Mathematics and Systems Science, Qingdao, 266590, China

[2] Shandong University of Science and Technology, College of Civil Engineering and Architecture, Qingdao, 266590, China

[3.] Qingdao Key Laboratory of Marine Civil Engineering Materials and Structures, Qingdao, 266590, China

## Abstract

Predicting the elevations of nonlinear wave fields behind floating breakwaters (FBs) is crucial for optimizing coastal engineering structures, enhancing safety, and improving design efficiency. Existing deep learning approaches exhibit limited generalization capability under unseen operating conditions. To address this challenge, this study proposes the Exogenous-to-Endogenous Frequency-Aware Network (E2E-FANet), a novel end-to-end neural network designed to model relationships between waves and structures. First, the Dual-Basis Frequency Mapping (DBFM) module leverages orthogonal cosine and sine bases to generate an adaptive time-frequency representation, enabling the model to effectively disentangle the evolving spectral components of wave signals. Second, the Exogenous-to-Endogenous Cross-Attention (E2ECA) module employs cross attention to explicitly model the unidirectional causal influence of floating breakwater motion on wave elevations. Additionally, a Temporal-wise Attention (TA) mechanism is incorporated that adaptively captures complex dependencies in endogenous variables. Extensive experiments, including generalization tests across diverse wave conditions and adaptability tests under varying relative water density (*RW*) conditions, demonstrate that E2E-FANet achieves superior predictive accuracy and robust generalization compared to mainstream models. This work emphasizes the importance of integrating causality and frequency-aware modeling in deep learning architectures for modeling nonlinear dynamics systems.

## 1. Introduction

The interaction between open-sea waves and offshore structures, particularly Floating Breakwaters (FBs), represents a classic problem in nonlinear dynamics. These complex systems, comprising a floating body and a mooring system, attenuate wave energy through a dynamic interplay of reflection, dissipation, and wave radiation [1]. However, with the intensification of climate change, mixed wave environments are becoming increasingly prevalent, characterized by the multi-scale coexistence of wind waves and long-period swells. In such scenarios, traditional linear superposition methods prove inadequate for wave behavior prediction, particularly when structures exhibit non-stationary responses to long-period swell forcing [2]. This limitation renders the wave elevations behind FBs highly unpredictable, thereby posing significant risks to port protection and operations in the lee of the breakwater. Consequently, developing models that can both accurately forecast wave elevations and reliably generalize to diverse sea conditions is not merely an engineering task, but a crucial challenge in nonlinear dynamics systems.

### 1.1 Literature review

Since the world's first FB was constructed in 1811, researchers have employed a hierarchy of methods to model the inherent nonlinear dynamics of this complex system. Traditional research approaches have primarily relied on physical experiments, theoretical analysis, and numerical simulations. For example, Physical model tests utilizing scaled prototypes have been conducted for both head and oblique waves, yielding data on structural motion response, mooring forces, and wave measurement characteristics [3-5]. Additionally, He et al. [6] used the method of characteristic function expansion to achieve continuity between the modified mild-slope equation and linear potential flow theory at the boundary, studying the hydrodynamic characteristics of a single-wing floating breakwater. However, theoretical approaches based on potential flow theory generally disregard fluid viscosity effects, thereby oversimplifying complex fluid dynamics. In contrast, numerical simulations based on Navier-Stokes equations can accurately model viscous effects and intricate nonlinear phenomena, such as wave breaking and vortex evolution, making them widely applicable for studying wave-structure interactions. Han and Dong [7] conducted numerical simulations coupling Smoothed Particle Hydrodynamics (SPH) and MoorDyn to investigate the hydrodynamic performance of intermediate-to-long-period waves interacting with a winged floating breakwater. Traditional methods focus on the impact of different structural forms, wave parameters, and structural size parameters on their motion response and wave attenuation performance [8-10]. However, the prohibitive computational demands of these methods render them impractical for real-time forecasting applications, particularly in dynamic marine environments where non-stationary effects predominate. This computational bottleneck represents a fundamental challenge in achieving efficient predictions of such complex system evolution.

To overcome the computational bottleneck of traditional methods, machine learning approaches have emerged as a compelling alternative for efficient and accurate waves prediction. These techniques can extract intricate patterns directly from observational data, circumventing the need for explicit physical modeling [11, 12]. Early pioneering efforts included autoregressive (AR) models [13], autoregressive moving average (ARMA) models [14], grey models [15], support vector machines (SVM) [16], and Gaussian process (GP) models [17, 18]. These investigations have demonstrated the significant potential of machine learning approaches for achieving efficient prediction capabilities [19]. Chen et al. [13] demonstrate that machine learning autonomously extracts features from data and optimizes parameters through backpropagation algorithms, enabling more adaptive and efficient predictions. Duan et al. [20] have demonstrated that machine learning techniques significantly enhance prediction accuracy while reducing computational costs, making real-time forecasting feasible. These pioneering machine learning approaches laid the groundwork for data-driven wave prediction. However, their inherent model capacity constraints limit their ability to capture the complex nonlinear and non-stationary dynamics characteristic of wave propagation and structural response, ultimately compromising prediction accuracy under complex sea conditions.

To more effectively capture non-stationary and nonlinear information hidden in the data, recurrent neural network (RNN) variants, particularly long short-term memory (LSTM) networks and gated recurrent units (GRU), have demonstrated significant advantages in nonlinear time series prediction [21-27]. Current research developments primarily follow two technical trajectories. The first focuses on spatio-temporal feature fusion, exemplified by CNN-LSTM hybrid architectures [28, 29]. These models implement a hierarchical processing scheme, utilizing convolutional neural networks for spatial feature extraction followed by recurrent neural networks for temporal relationship modeling. The second trajectory emphasizes signal decomposition techniques, incorporating methods such as Fourier Transform (FT), Empirical Mode Decomposition (EMD) and Variational Mode Decomposition (VMD) in the preprocessing phase [30-33]. While these methods have significantly advanced the field, their architecture often exhibit inherent limitations in capturing the long-range dependencies and multi-scale interactions characteristic of the wave dynamics described previously, thereby constraining their generalization capabilities [34]. The Transformer architecture, through its self-attention (SA) mechanism, offers an effective solution for capturing long-range dependencies in time series, demonstrating superior performance over traditional RNN models across multiple benchmark tests [35]. These models excel at modeling long-term patterns and periodic features by integrating information from arbitrary positions within sequences. Song et al. [36] proposed the spatio-temporal-variable transformer (STVformer), which enables simultaneous extraction of long-term and multi-scale dynamic features while capturing significant correlations. Wu et al. [37] proposed the Autoformer architecture, which leverages FT to compute time series autocorrelation, leading to substantial improvements in both computational efficiency and prediction accuracy.

The aforementioned literature review demonstrates significant advances in nonlinear time series prediction research. However, when applied to nonlinear dynamics systems such as FBs, existing deep learning models exhibit limited generalization capabilities. This limitation stems from two major technical challenges. First, traditional self-attention mechanisms indiscriminately mix all input variables, failing to capture the fundamental physical causal relationships within the system. Specifically, they neglect the directional influence of FB motion (exogenous variables) on wave field changes (endogenous variables), resulting in models that learn spurious statistical correlations rather than robust physical relationships. Second, these approaches rely on static frequency analysis, which is inadequate for capturing the dynamics inherent in wave field evolution. This constrains the model's ability to adaptively capture the spectral characteristics of wave field variations, thereby limiting generalization capabilities under varying sea conditions.

## 1.2 Motivation and contributions

Addressing these research challenges, this study proposes an Exogenous-to-Endogenous Frequency-Aware Network (E2E-FANet). By resolving the two core deficiencies in existing deep learning models, E2E-FANet achieves a more robust understanding of the system's nonlinear dynamics. First, we propose a Dual-Basis Frequency Mapping (DBFM) module that supersedes static frequency analysis with learnable mapping, thereby enabling adaptive capture of fundamental spectral characteristics inherent in wave dynamics. Second, we developed an Exogenous-to-Endogenous Cross-Attention mechanism (E2ECA) that explicitly models the physical causal relationships from structural motion to wave field, enabling the model to learn directional physical interaction patterns and enhance robustness. Through the synergistic integration of DBFM and E2ECA modules, E2E-FANet achieves physics-guided modeling of wave-structure interactions, significantly improving prediction accuracy and generalization capability.

The main contributions of this study can be summarized as follows:

(1) We propose E2E-FANet, a novel deep learning framework specifically designed to enhance the generalization capability of wave elevation behind FBs.
(2) We develop an E2ECA mechanism that enables explicit modeling of causal relationships between floating breakwater motion and wave elevations. This enables the model to learn interaction patterns with clear physical meaning while avoiding spurious correlations.
(3) We introduce the DBFM module, which employs orthogonal cosine and sine bases to capture fundamental time-frequency domain features, enabling the model to achieve better generalization across diverse wave conditions.
(4) We conduct comprehensive experiments on wave flume datasets with varying wave conditions, demonstrating that our method outperforms baseline model in terms of prediction accuracy and generalization capability. Further experiments on cross-scenario generalization demonstrate E2E-FANet's robust adaptability to different breakwater systems, validating its practical applicability for early warning systems and harbor safety management in dynamic marine environments.

The remainder of this study is structured as follows: Section 2 details the proposed E2E-FANet framework. Section 3 describes the main experimental methodology and results. Section 4 presents generalization experiment analysis across different wave conditions. Section 5 presents adaptability testing with varying relative water density. Section 6 concludes the study with key findings.

# 2.Model

## 2.1 Problem definition

Given an endogenous time series $X=\{\mathbf{x}_1,\mathbf{x}_2,\ldots,\mathbf{x}_L\}\in\mathbb{R}^{L\times F}$ and the corresponding exogenous sequence $Z=\{\mathbf{z}_1,\mathbf{z}_2,\ldots,\mathbf{z}_L\}\in\mathbb{R}^{L\times C}$, where $L$ is the look back window size, $F$ is the number of endogenous variables, and $C$ is the number of exogenous variables. In this study, the exogenous variables comprise three degrees of freedom motion responses of the floating body: heave, surge, and pitch. These motion responses, although induced by upstream waves, are classified as exogenous factors since they influence downstream waves while remaining unaffected by downstream wave feedback within the prediction time. The endogenous variables consist of wave elevation measured at various positions both upstream and downstream of the floating breakwater. These wave elevations are classified as endogenous because they are influenced by both their historical values and the exogenous variables.

## 2.2 Framework overview

To enhance the generalization capability for predicting complex wave elevations behind FB, we propose an end-to-end framework termed E2E-FANet. As illustrated in Fig. 1, the overall architecture consists of four core modules: (1) Embedding, a dual-component module that transforms raw measurements into representation space while preserving temporal information; (2) Dual-Basis Frequency Mapping Module, a module that employs learnable orthogonal cosine and sine basis functions to capture fundamental frequency domain features; (3) Temporal-wise Attention, a mechanism that adaptively weights different time steps to focus on critical temporal patterns; and (4) Exogenous-to-Endogenous Cross-Attention, a specialized attention mechanism designed to explicitly model the physical causality from the breakwater's motion to the wave elevations. These modules work synergistically to enhance waves prediction accuracy behind floating breakwater, particularly in complex scenarios where traditional methods exhibit limitations.

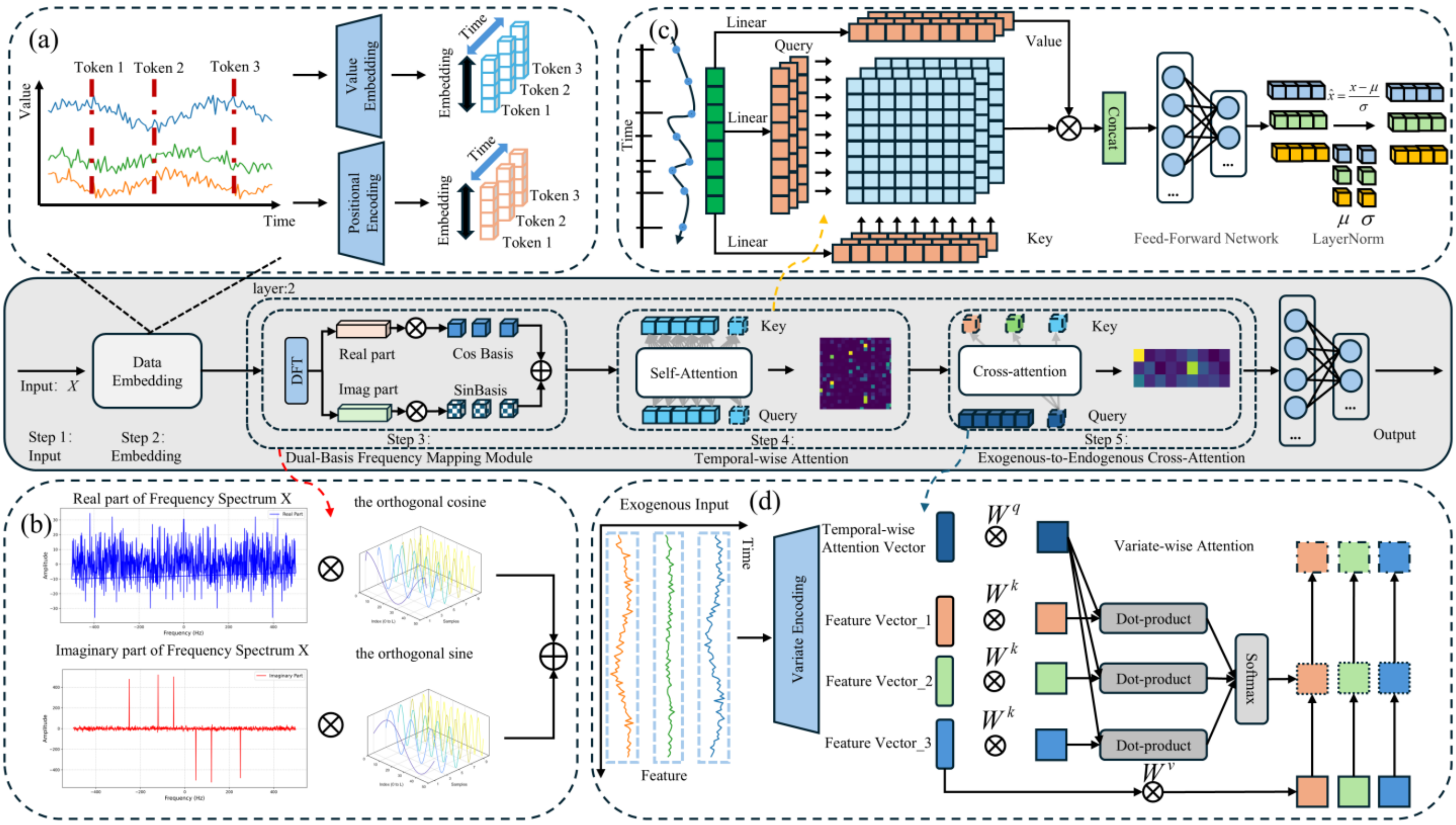

Fig. 1. Framework of the proposed E2E-FANet.

## 2.3 Data embeddings

The Transformer-based model initially converts input sequences into vector representations through an embedding layer. We implement a dual component embedding strategy comprising Value Embedding (VE) and Position Encoding (PE), as illustrated in Fig. 1(a). The VE component transforms raw input features at each time step into a continuous vector space. This transformation converts discrete input features into dense vector representations, making them more suitable for neural network processing. Specifically, for an endogenous variable sequence $X=\{\mathbf{x}_1,\mathbf{x}_2,\ldots,\mathbf{x}_L\}\in\mathbb{R}^{L\times F}$, where $\mathbf{x}_t=\left[x_{t,1},x_{t,2},\ldots,x_{t,F}\right]$ represents the vector at time step $t$, $L$ denotes the look back window size, and $F$ represents the number of endogenous variates. The VE component at time step $t$ is computed as:

$$VE(\mathbf{x}_t) = \mathbf{x}_t \mathbf{E} \in \mathbb{R}^D \tag{1}$$

where $\mathbf{E} \in \mathbb{R}^{F \times D}$ is the embedding matrix. This embedding matrix is shared across all time steps and features, allowing the model to learn a consistent mapping from the input feature space to a $D$-dimensional embedding space. $D$ is a hyperparameter that controls the capacity of the embedding space to represent feature information.

To incorporate temporal information into the model, we implement sinusoidal position encoding, a widely adopted and effective technique, to inject information about the position of each time step within the input sequence. This approach encodes temporal information using sinusoidal functions of different frequencies. For a given time step $t$ and embedding dimension $D$, the PE is calculated as follows:

$$\begin{aligned} PE(t,2i) &= \sin\left(\frac{t}{10000^{\frac{2i}{D}}}\right) \\ PE(t,2i+1) &= \cos\left(\frac{t}{10000^{\frac{2i}{D}}}\right) \end{aligned} \tag{2}$$

where $t$ denotes the temporal index, $i$ represents the dimension index within the input vector at each position, and $D$ is the embedding dimension. The constant 10000 follows Vaswani et al. [38] to ensure positional distinguishability across long sequences.

Finally, the dual-component embedding for each time step $t$, denoted as **E(t)**, is obtained by element-wise addition of the VE and PE components:

$$\mathbf{E}(t) = VE(\mathbf{x}_t) + PE(t) \tag{3}$$

The element-wise addition combines feature-specific information from VE with temporal information from PE, creating a comprehensive representation of the input sequence for subsequent layers of E2E-FANet. This integrated embedding strategy ensures the model captures both the input feature content and their temporal sequence, essential for accurate waves prediction.

## 2.4 Dual-basis frequency mapping module

Wave phenomena are inherently characterized by their frequency content. Traditional approaches, such as directly applying DFT, often utilize frequency coefficients for analysis. However, this direct approach may lead to imprecise interpretation of frequency coefficients and overlook the fundamental basis functions that constitute the frequency spectrum [39]. To learn fundamental spectral features of wave elevations, we introduce the DBFM module. DBFM employs learnable orthogonal sine and cosine basis functions to parametrically decompose input sequences in the spectral domain. Unlike the static transformation of DFT, DBFM adaptively learns to extract task-relevant frequency components while maintaining local temporal characteristics.

The primary strength of DBFM lies in constructing frequency features, as illustrated in Fig. 1(b). DBFM module leverages the orthogonal nature of cosine and sine functions, which form the real and imaginary parts of the complex exponential basis functions in DFT. These orthogonal bases ensure linear independence, allowing for the capture of distinct and non-redundant frequency components within the signal. The process begins with generating the frequency spectrum **H** of the input embedding sequence using the DFT:

$$\mathbf{H}(k) = DFT(\mathbf{E}) = \sum_{n=0}^{L-1} \mathbf{E}[n] \exp\left(-i\frac{2\pi kn}{L}\right), \quad k = 0,1,\ldots,L-1 \tag{4}$$

where $\mathbf{H}(k)$ is a complex-valued frequency-domain signal derived from the input embedding $\mathbf{E}$, $L$ represents the look back window size, and $\exp(\bullet)$ denotes the complex exponential basis function at frequency $k$. The DFT essentially decomposes the time-domain signal $\mathbf{E}$ into a sum of complex exponentials at different frequencies.

We define the orthogonal cosine basis (CosBasis) and sine basis (SinBasis) as follows:

$$\begin{aligned}\text{CosBasis} &= \frac{1}{L}\left[1, 2\cos\left(\frac{2\pi\mathbf{N}}{L}\right), \ldots, 2\cos\left(\frac{(L-1)\pi\mathbf{N}}{L}\right), \cos(\pi\mathbf{N})\right] \\ \text{SinBasis} &= -\frac{1}{L}\left[0, 2\sin\left(\frac{2\pi\mathbf{N}}{L}\right), \ldots, 2\sin\left(\frac{(L-1)\pi\mathbf{N}}{L}\right), \sin(\pi\mathbf{N})\right]\end{aligned} \tag{5}$$

More intuitively, in this formula, $\mathbf{N} = [0, 1, \ldots, L-1]$ represents a series of time points or index values. When computing the cosine and sine bases, we need to generate two matrices, CosBasis and SinBasis, rather than single values. Here, CosBasis and SinBasis are $L \times L$ matrices, where each row corresponds to a basis function at a specific frequency, and each column corresponds to a time point. Specifically, for the $k$-th row of the CosBasis and SinBasis matrices, the formulas are as follows:

$$\begin{aligned}\text{CosBasis}(k) &= \frac{1}{L}\left[1, 2\cos\left(\frac{2\pi k}{L}\right), \ldots, 2\cos\left(\frac{(L-1)\pi k}{L}\right), \cos(\pi k)\right] \\ \text{SinBasis}(k) &= -\frac{1}{L}\left[0, 2\sin\left(\frac{2\pi k}{L}\right), \ldots, 2\sin\left(\frac{(L-1)\pi k}{L}\right), \sin(\pi k)\right]\end{aligned} \tag{6}$$

The incorporation of both cosine and sine bases enables DBFM to represent the complete frequency spectrum, encompassing both real and imaginary components, which effectively characterizes the amplitude and phase information. By multiplying the real part of $\mathbf{H}$ (denoted as $\mathbf{H_R}$) with the orthogonal cosine basis and the imaginary part of $\mathbf{H}$ (denoted as $\mathbf{H_I}$) with the orthogonal sine basis, a mixed representation of frequency and temporal features is obtained as follows:

$$\begin{aligned}\mathbf{F_R} &= \mathbf{H_R} \cdot \text{CosBasis} \\ \mathbf{F_I} &= \mathbf{H_I} \cdot \text{SinBasis}\end{aligned} \tag{7}$$

Here, $\cdot$ denotes element-wise multiplication. $\mathbf{F_R}$ and $\mathbf{F_I}$ represent the cosine-basis and sine-basis frequency features, respectively. The frequency features are generated through the integration of DFT frequency components with their corresponding basis functions, yielding a more comprehensive and interpretable frequency representation than conventional approaches using raw DFT coefficients alone. Due to the Hermitian symmetry property of the Fourier Transform, computing only the first $L/2+1$ frequency components sufficiently represent the complete frequency-domain information. Finally, we concatenate the cosine and sine frequency features horizontally to form the frequency-domain representation matrix $\mathbf{G}$: $\mathbf{G} = \mathbf{F}_R \oplus \mathbf{F}_I$, where $\oplus$ denotes horizontal concatenation. This DBFM procedure effectively decomposes the original time series into multiple frequency-level components $\mathbf{G}$ while maintaining the complete temporal structure. This enables the model to not only capture core frequency components but also preserve time-varying characteristics crucial for prediction tasks, thereby achieving more accurate wave prediction.

## 2.5 Temporal-wise attention

Traditional RNNs, while designed for sequential data, can struggle with capturing long-range temporal dependencies and may suffer from vanishing or exploding gradients. Attention mechanisms, particularly self-attention, offer a powerful alternative to directly address relevant time steps across the entire input sequence, regardless of their temporal distance. To achieve this, we integrate TA mechanism into E2E-FANet.

As shown in Fig.1(c), TA mechanism is implemented using multi-head attention. For each head $h$, the query $Q_h$, key $K_h$, and value $V_h$ are computed as:

$$Q_h = X^T W_h^Q, K_h = X^T W_h^K, V_h = X^T W_h^V \tag{8}$$

where $W_h^Q, W_h^K, W_h^V \in \mathbb{R}^{D\times d}$ are learnable parameters. These projections transform the input embedding into $Q_h$, $K_h$, and $V_h$ representations that are used to compute attention weights. The dot-product attention operation for each head is computed as:

$$\text{Attention}\left(Q_h, K_h, V_h\right) = \text{Softmax}\left(Q_h K_h^T\right) V_h \tag{9}$$

The attention weights, calculated by the Softmax function, determine the importance of each time step in the input sequence when computing the weighted sum of values. The final output of the temporal-wise attention mechanism is obtained by concatenating the outputs of all attention heads and then linearly transforming the concatenated result:

$$\begin{aligned} &\text{head}_h = \text{Attention}\left(Q_h, K_h, V_h\right), \\ &\text{TWA}(X) = \text{Concat}\left(\text{head}_1, \ldots, \text{head}_h\right)^T W_O \end{aligned} \tag{10}$$

where $W_O \in \mathbb{R}^{hd\times D}$ are learnable parameters. To further improve convergence and training stability, we apply layer normalization to the output of the multi-head attention:

$$E = \text{LN}\left(X + \text{TWA}(X)\right) \tag{11}$$

The temporal-wise attention mechanism enables E2E-FANet to selectively identify and focus on temporal patterns within wave data. This approach enhances model capability for capturing complex temporal dependencies.

## 2.6 Exogenous-to-endogenous cross-attention

To effectively model the influence of exogenous factors on endogenous variables, we incorporate E2ECA component within E2E-FANet, comprising variate embedding and variate-wise cross-attention mechanisms.

### 2.6.1 Variate embedding

Conventional Transformer-based forecasting models often treat multivariate time series as a single input sequence, embedding all variables together as temporal tokens. However, this approach may not effectively capture the distinct characteristics of individual variables and their specific inter-relationships [40]. To implement variate-wise processing, we first employ variate embedding layers to encode each individual time series as a distinct variate token. This contrasts with traditional temporal tokenization where all variables at a given time step are embedded together. Specifically, we use separate linear projection layers for endogenous and exogenous variables:

$$\begin{aligned} &V_{en} = \text{EnVariateEmbed}(E): \mathbb{R}^L \to \mathbb{R}^D, \\ &V_{ex} = \text{ExVariateEmbed}\left(Z\right): \mathbb{R}^L \to \mathbb{R}^D. \end{aligned} \tag{12}$$

where EnVariateEmbed and ExVariateEmbed are trainable linear projection layers encoding the endogenous variable and exogenous variable, respectively. The parameters $L$ represents the sequence lengths of endogenous and exogenous variables, while $D$ denotes the target dimensionality of the encoded feature space. $V_{en}$ and $V_{ex}$ are the resulting variate embeddings for endogenous and exogenous variables. By using separate embedding layers, E2E-FANet learn distinct representations for endogenous and exogenous variables, capturing their unique characteristics.

### 2.6.2 Variate-wise cross-attention

To explicitly model the influence of exogenous variables on endogenous variables, we employ a variate-wise cross-attention (VCA). Drawing inspiration from multi-modal learning applications [41], we utilize cross-attention to capture dependencies between the variate embeddings. In this cross-attention layer, VCA treats endogenous variable embeddings $V_{en}$ as queries and exogenous variable embeddings $V_{ex}$ as both keys and values. This directional attention flow aligns with the fundamental physical principle that exogenous variables influence endogenous wave dynamics. The query $Q_h$, representing endogenous wave elevations, attends to relevant information from the keys $K_h$ and values $V_h$ to capture how the motion of the breakwater influences downstream wave characteristics. This approach represents a significant departure from standard self-attention mechanisms, which process all variables symmetrically. Specifically, for each attention head *h*:

$$Q_h = V_{en}^T W_h^Q, K_h = V_{ex}^T W_h^K, V_h = V_{ex}^T W_h^V, \tag{13}$$

where $W_h^Q, W_h^K, W_h^V \in \mathbb{R}^{D\times d}$ are learnable parameters. $V_{en}$ is endogenous variable embedding, and $V_{ex}$ is exogenous variable embedding. The dot-product cross-attention computation follows the standard attention formulation:

$$\text{CrossAttention}(Q_h, K_h, V_h) = \text{Softmax}(Q_h K_h^T) V_h \tag{14}$$

This cross-attention mechanism enables the model to learn attention weights that quantify the relevance influence of each exogenous variable on each endogenous variable. The process facilitates information transfer from exogenous variable tokens to endogenous variable tokens, enabling propagation of learned multivariate dependencies and enhancing model ability to capture the influence of external factors on waves prediction.

The outputs of all cross-attention heads are concatenated, and then a linear transformation is applied to the concatenated result to obtain the final output:

$$\begin{aligned} &\text{head}_h = \text{CrossAttention}(Q_h, K_h, V_h), \\ &\text{VCA} = \text{Concat}(\text{head}_1, \ldots, \text{head}_h)^T W_O, \\ &\hat{y}_i = \text{Linear}(\text{VCA}(X)) \end{aligned} \tag{15}$$

2.7 Loss function

The prediction accuracy is quantified using the mean squared error (MSE) loss function, formulated as:

$$Loss = \frac{1}{n}\sum_{i=1}^{n}(y_i - \hat{y}_i)^2 \tag{16}$$

where $y_i$ is the measured value for the *i-th* data point, $\hat{y}_i$ is the predicted value for the *i-th* data point generated by the model.

## 3. Experimental results and analysis

### 3.1. Datasets details

To train and evaluate the proposed E2E-FANet for waves prediction behind FB, we utilized a dataset generated from a numerical wave flume simulation. This section details the data generation process, wave conditions, and dataset characteristics. In this study, MoorDyn and DualSPHysics coupling model are used to realize the interaction between waves and FB. The interaction between fluid and floating body is simulated by DualSPHysics, and the angular and linear velocities of floating body are obtained. These are imported with the SPH time step into MoorDyn to update the position and velocity of the fairleads. Then, the forces of the concentrated mass point on the mooring lines are calculated. Finally, the velocity and position of the floating body are updated. More details are shown in Appendix A. The numerical model is verified in Han and Dong [7]. The size of box-type floating breakwater is 0.5×0.2 m, and the density is 500 kg/m$^3$. The length of each mooring line is 1.58 m, and the weight per unit length is 0.06 kg/m. The numerical simulation was conducted for a duration of 500 s, with a sampling interval of 0.05 s. The arrangement of wave flume and floating breakwater is shown in Fig. 2. There are five wave gauges set in front of the floating breakwater, and four wave gauges are behind. The wave conditions are irregular waves, as shown in Table 1. To reduce the wave secondary reflection, relaxation wave making zone is adopted and the wave damping zone is set at the two ends of the flume. Therefore, the dataset includes five wave gauges in front of FB, the motion response of FB, four wave gauges behind FB. The input variables include wave elevation measurements from five upstream wave gauges (WG1-WG5), and three floating breakwater motion responses (surge, heave, and pitch). The prediction target consists of wave elevation measurements from four downstream wave gauges (WG6-WG9), which record wave conditions behind the breakwater after transformation. E2E-FANet aims to predict these downstream wave elevations using upstream wave measurements and breakwater motion responses as inputs. For model training and evaluation, the continuous time series data were segmented into sequences with a fixed input window length of 48-time steps. Each sequence's prediction target comprised the wave elevations over the subsequent 48-time steps. The dataset was chronologically partitioned into training (70%), validation (10%), and test sets (20%).

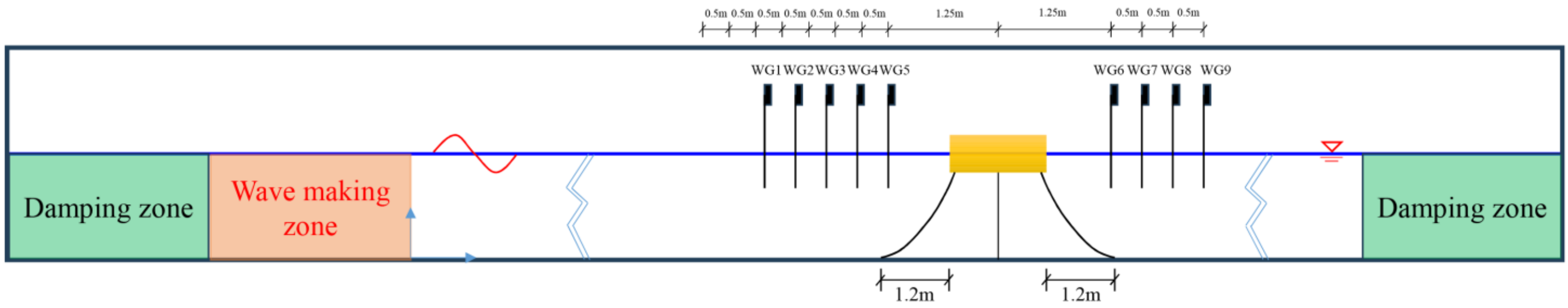

Fig. 2. Schematic diagram of the wave flume and floating breakwater interaction in numerical simulation.

Table 1 Wave conditions used in numerical wave flume simulations.

| Water depth (m) | Peak wave period $T_p$ (s) | Significant wave elevations $H_s$ (m) |
|---|---|---|
| 0.8 | 1.0, 1.5, 2.0, 2.5 | 0.12, 0.15, 0.18, 0.21 |

## 3.2. Evaluation metrics

To quantitatively evaluate the performance of the E2E-FANet and the baseline models in predicting wave elevations behind the breakwater, we employed four standard statistical metrics. These metrics are widely used in regression tasks and provide a comprehensive assessment of prediction accuracy and error characteristics. The chosen metrics are Mean Squared Error (MSE), Mean Absolute Error (MAE), Root Mean Square Error (RMSE), and Mean Absolute Percentage Error (MAPE). These metrics are defined as follows:

$$MSE = \frac{1}{n}\sum_{i=1}^{n}\left(\hat{y}_i - y_i\right)^2 \tag{17}$$

$$MAE = \frac{1}{n}\sum_{i=1}^{n}\left|\hat{y}_i - y_i\right| \tag{18}$$

$$RMSE = \sqrt{\frac{1}{n}\sum_{i=1}^{n}\left(\hat{y}_i - y_i\right)^2} \tag{19}$$

$$MAPE = \frac{100\%}{n}\sum_{i=1}^{n}\left|\frac{\hat{y}_i - y_i}{y_i}\right| \tag{20}$$

where $n$ represents the total number of the predicted samples, $y_i$ represents the measured values, and $\hat{y}_i$ represents the predicted values.

## 3.3 Baseline models and experimental settings

To evaluate the E2E-FANet model's performance comprehensively, we selected baseline models from three distinct categories of time series forecasting approaches relevant to wave prediction: sequential models, frequency-domain models, and Transformer-based models. Each category offers unique capabilities in capturing complex patterns in time series data. The first category comprises sequential models, which are widely used for time series analysis due to their ability to process sequential data and capture temporal dependencies. This category includes Long Short-Term Memory (LSTM) networks [42],

Temporal Convolutional Network (TCN) [43], and the integrated CNN-LSTM architecture. The second category includes frequency-domain models, specifically FiLM [44] and FreTS [45], which utilize Fourier transforms and related spectral analysis techniques. Including frequency-domain baselines enables evaluation of whether E2E-FANet's integration of frequency-domain analysis provides advantages over purely frequency-domain-based approaches. The third category comprises Transformer-based architectures, including Crossformer [46], iTransformer [47], and FEDformer [48], which leverage attention mechanisms to capture long-range temporal dependencies. These baseline models, consisting of the original Transformer and its advanced variants, represent state-of-the-art achievements in attention-based sequence modeling for time series forecasting.

All models were implemented using the PyTorch framework and optimized using the Adam optimization algorithm. The hyperparameters were configured as follows: initial learning rate of 1e-3, mini-batch size of 32, and maximum of 20 training epochs. To prevent overfitting, two regularization strategies were employed: dropout with a rate of 0.1 in the output layers, and early stopping with a patience of three epochs when the validation loss showed no improvement. All experiments were conducted on a single NVIDIA GeForce RTX 3090 GPU.

The validity and stability of our experimental results were monitored by tracking the loss function values for both training and validation sets throughout the training epochs. Fig. 3 presents the training loss curves for E2E-FANet and all baseline models, illustrating their training stability and convergence trends. Each subplot corresponds to a specific model, illustrating the progression of training and validation losses across epochs. The close tracking between training and validation loss curves for most models, with no significant upward divergence in validation loss after reaching its minimum, indicates effective control of overfitting. The implementation of dropout regularization and early stopping mechanism successfully prevented excessive memorization of training data while enhancing generalization capability. The loss curves in Fig. 3 validate the effectiveness and stability of the experimental settings and training procedures.

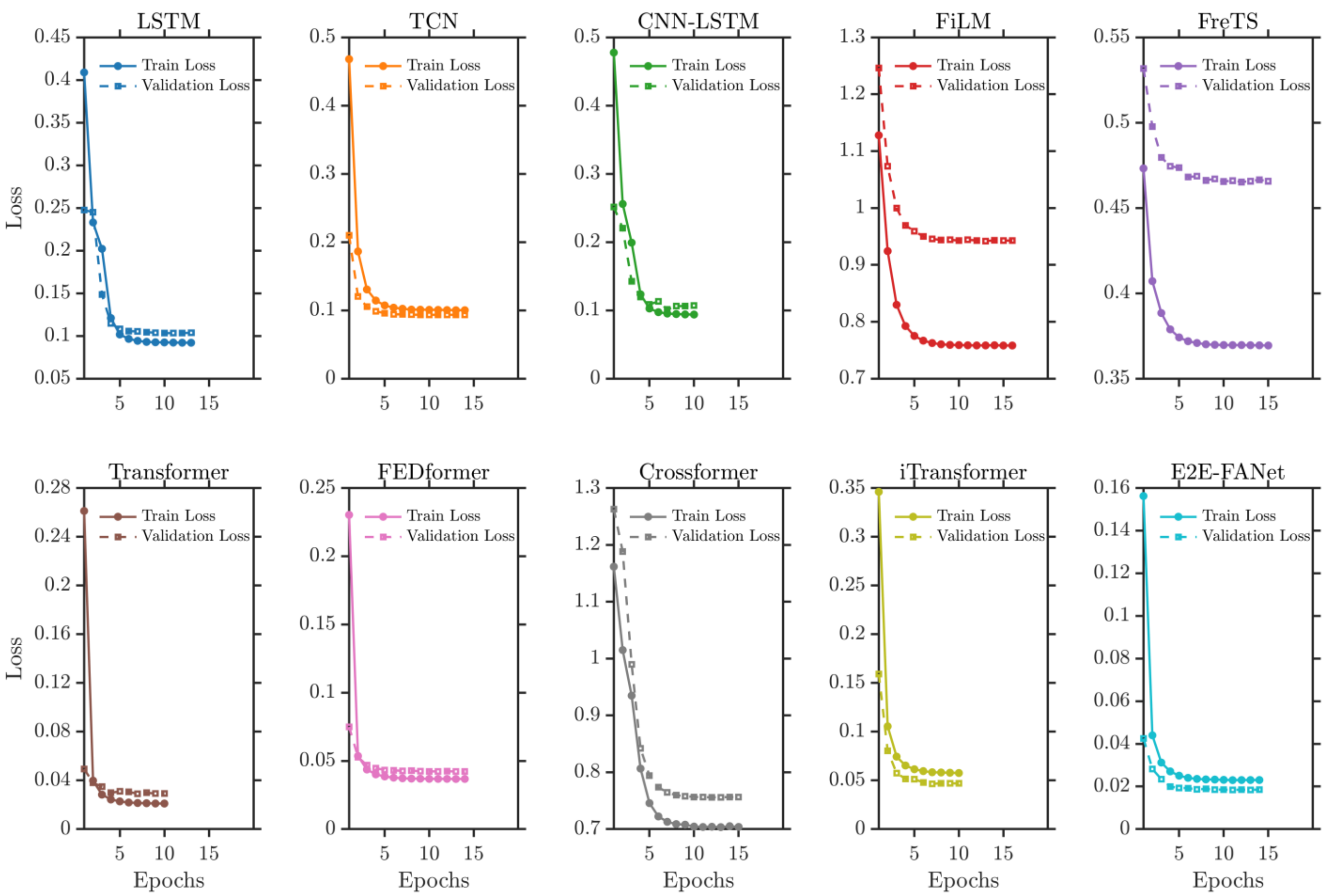

Fig. 3. Training and validation loss curves for E2E-FANet and baseline models.

## 3.4 Overall performance analysis

### 3.4.1 Main results and analysis

Table 2 presents a quantitative comparison of E2E-FANet with LSTM, TCN, CNN-LSTM, FiLM, FreTS, Transformer, FEDformer, Crossformer, and iTransformer on the dataset with $H_s$ = 0.18 m, $T_p$ = 2 s. The best and second-best results are highlighted in bold and underlined, respectively. The table clearly shows that E2E-FANet achieves superior performance across all evaluation metrics, with significant improvements over the second-best baseline model. Specifically, MSE, RMSE, and MAE are reduced by 19.95%, 10.64%, and 12.88%, respectively, and MAPE is reduced by 2.74%. These significant quantitative improvements demonstrate E2E-FANet's effectiveness in enhancing waves prediction accuracy.

Analysis of the baseline model performance reveals distinct patterns across different architectural categories. RNN-based architectures (LSTM, TCN, and CNN-LSTM) show low prediction accuracy due to their shallow architectures. Their sequential processing constrains effective modeling of complex interactions between upstream wave measurements and floating breakwater motion responses. Frequency-domain models (FiLM and FreTS) incorporate frequency-domain processing mechanisms, yet their performance indicates that purely frequency-domain modeling is insufficient for this complex, multi-factor time-varying system. Transformer-based models demonstrate strong overall performance, with the standard Transformer approaching optimal results through effective self-attention mechanisms. Although Transformer variants such as iTransformer and Crossformer are designed to capture multivariate correlations, they exhibit lower efficiency in capturing variable relationships essential for waves prediction, performing below the standard Transformer model. The proposed E2ECA mechanism of E2E-FANet, which approaches the problem from the perspective of exogenous-to-endogenous interactions, more effectively captures these critical relationships for wave prediction, demonstrating its advantages in addressing such challenges.

Fig. 4 illustrates scatter density plots comparing E2E-FANet and baseline model performances, with measured values on the horizontal axis and model predictions on the vertical axis. The ideal prediction line (y=x) is indicated by a dashed black line, while the actual regression fit is shown in red. The color gradient from blue to red indicates the density of sample points. The results demonstrate that E2E-FANet predictions closely align with the ideal line, with the regression fit nearly coinciding with y=x, indicating minimal prediction errors and high model stability. Density analysis reveals that E2E-FANet achieves particularly strong clustering in high-density regions. In contrast, baseline models display greater scatter, particularly in low-density regions where long-tail effects are prominent, indicating limited capability in predicting extreme events. These results demonstrate E2E-FANet's effectiveness in predicting wave elevations behind FB emphasizing its hybrid architecture's ability to capture both frequency-domain features and cross-variable interactions.

Table 2 Comparison of model performance with $H_s$ = 0.18 m, $T_p$ = 2 s

| Model type | Model | MSE | MAE | RMSE | MAPE |
|---|---|---|---|---|---|
| RNN-based methods | LSTM | 0.1201 | 0.2624 | 0.3466 | 2.0664 |
| | TCN | 0.1480 | 0.2912 | 0.3847 | 2.2087 |
| | CNN-LSTM | 0.1228 | 0.2690 | 0.3505 | 2.2402 |
| Fourier-based methods | FiLM | 0.9042 | 0.7425 | 0.9509 | 3.8401 |
| | FreTS | 0.5366 | 0.5390 | 0.7326 | 2.7007 |
| Transformer-based methods | Transformer | <u>0.0431</u> | <u>0.1561</u> | <u>0.2078</u> | <u>1.1426</u> |
| | FEDformer | 0.0510 | 0.1657 | 0.2260 | 1.3250 |
| | Crossformer | 0.8525 | 0.7010 | 0.9233 | 3.0502 |
| | iTransformer | 0.0472 | 0.1603 | 0.2173 | 1.1838 |
| | E2E-FANet | **0.0345** | **0.1360** | **0.1857** | **1.1113** |

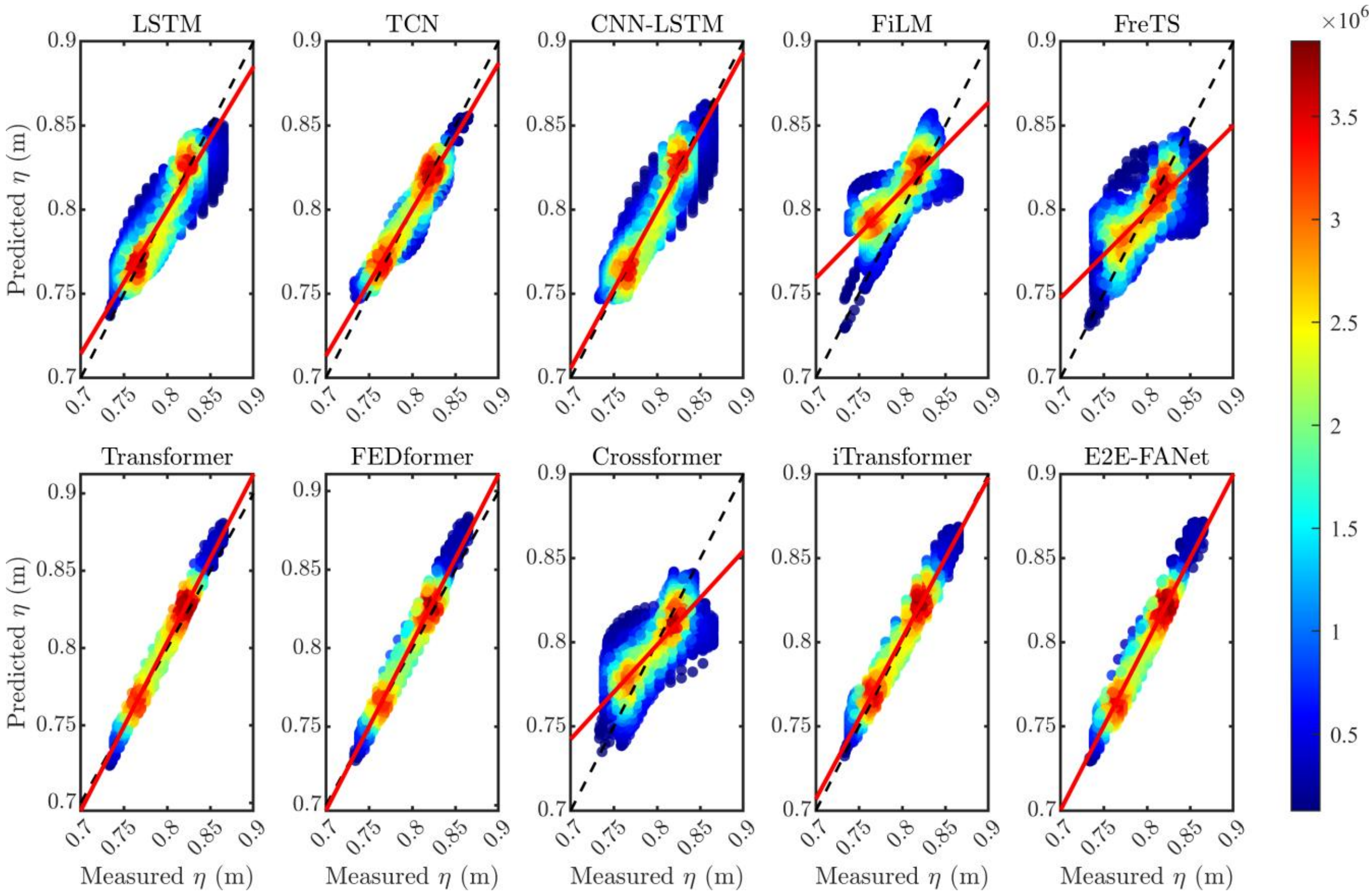

Fig. 4. Scatter density plots of E2E-FANet and baseline models with $H_s$ = 0.18 m, $T_p$ = 2 s.

3.4.2 Performance analysis at each downstream wave gauge

This section examines prediction performance at individual downstream wave gauges (WG6-WG9) through visual and quantitative assessment. Fig. 5 presents time-series comparisons between model predictions and measured wave elevations at each gauge location, while Fig. 6 quantifies the cumulative absolute error for each model at each wave gauge, enabling the assessment of prediction stability across various locations behind the breakwater. As shown in Fig. 5, E2E-FANet trajectories maintain the closest proximity to the measured values, demonstrating high accuracy in capturing temporal wave elevations dynamics at each location. While other models capture general wave patterns, they exhibit notable deviations from measured values, particularly at wave extrema. LSTM and TCN tend to underestimate wave crests in WG6 and WG8, while the Transformer exhibits a tendency to overestimate the peaks, which is most pronounced at WG7 and WG9. The bar charts in Fig. 6 present the cumulative absolute error for each model at each wave gauge, providing a comprehensive measure of total prediction error over the time series. E2E-FANet achieves the lowest cumulative absolute error across all wave gauges. The error magnitudes for E2E-FANet show minimal variation across locations, demonstrating both superior accuracy and stability. Baseline models exhibit higher cumulative errors. The integrated visual and quantitative analysis demonstrates that E2E-FANet exhibits superior performance at the downstream wave gauges, with the lowest prediction errors across all locations and minimal variation.

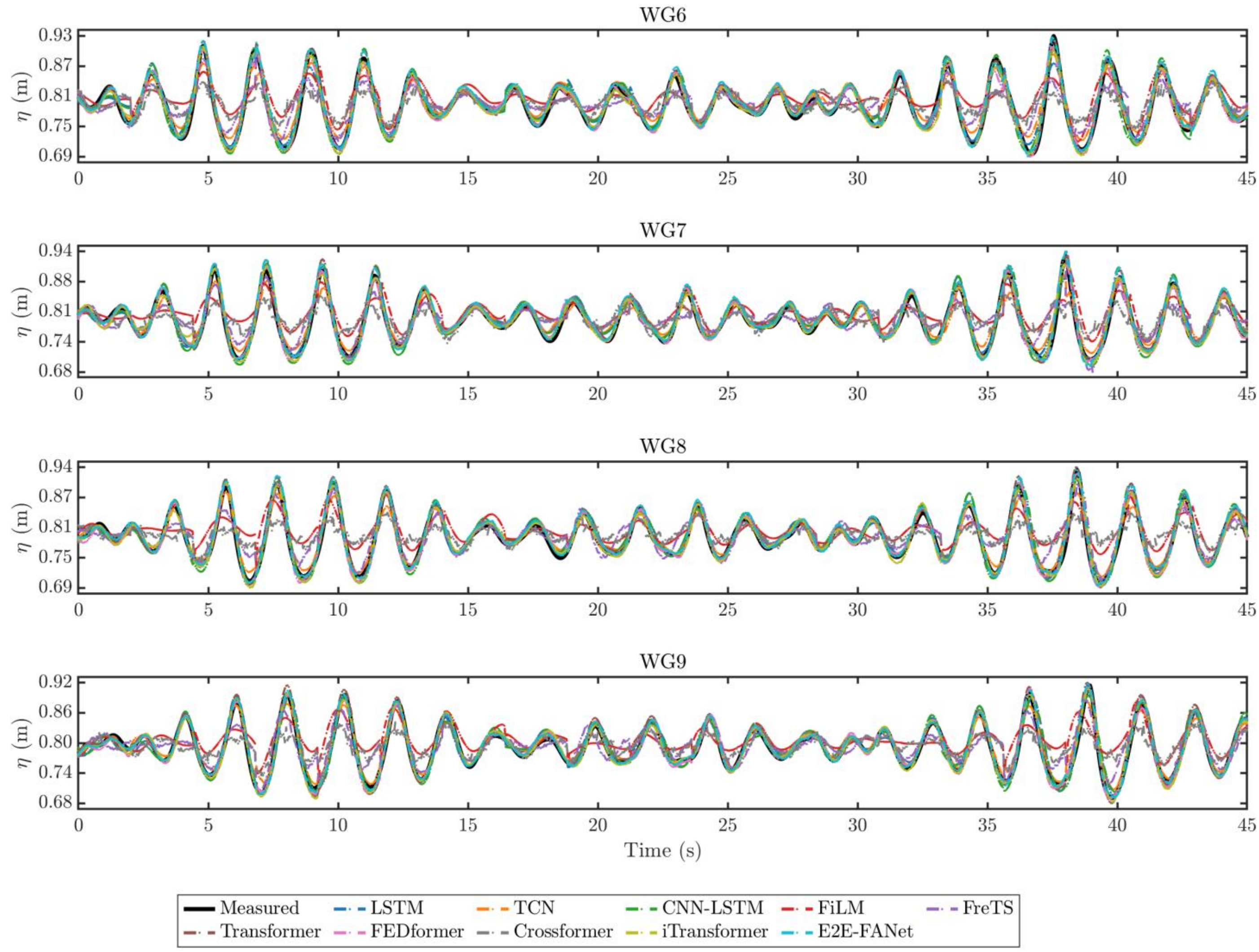

Fig. 5. Comparison of predicted and measured $H_s$ for E2E-FANet and baseline models at downstream wave gauges (WG6-WG9).

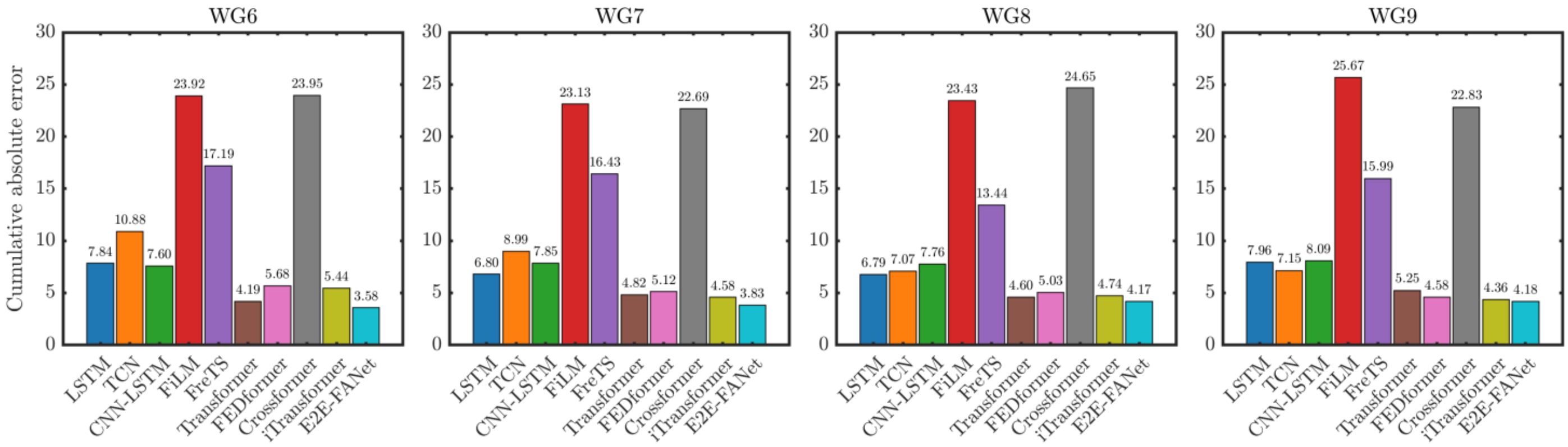

Fig. 6. Cumulative absolute error of E2E-FANet and baseline models at downstream wave gauges (WG6-WG9).

3.4.3 Impact of different prediction horizons

To evaluate model performance across different temporal scales, we conducted forecasting experiments with a 48-step horizon, analyzing predictions at eight representative time points: $t$+1, $t$+7, $t$+13, $t$+19, $t$+25, $t$+31, $t$+37, and $t$+43 (in 6-step increments). This systematic approach includes short-term, mid-term, and long-term forecasts, enabling comprehensive performance assessment across temporal scales. Fig. 7 presents scatter density distributions for all models at these forecast horizons, with columns representing different models and rows indicating forecast steps. The measured values are plotted on the x-axis against model predictions on the y-axis, allowing visual assessment of prediction accuracy and stability through the proximity of scatter points to the ideal line and their density distributions. The results reveal different patterns in model performance across different time horizons. At shorter intervals (*t*+1, *t*+7), most models demonstrate strong predictive capability, evidenced by tight clustering around the ideal line. However, performance differentiation becomes increasingly apparent as the forecast horizon extends. Transformer maintains relatively concentrated scatter distributions for most horizons, they exhibit increasing outliers at extended lead times, suggesting reduced effectiveness in long-range forecasting. The increasing outlier occurrence in Transformer at longer horizons might indicate that their attention mechanisms, while powerful, may become less focused or more prone to capturing noise as the prediction horizon expands, leading to error accumulation. iTransformer, FEDformer show strong performance in short and mid-range predictions but demonstrates increased scatter at longer horizons. Crossformer exhibits more dispersed distributions across all timescales, with pronounced tail effects and outlier clusters. In contrast, E2E-FANet maintains consistent predictive accuracy across all time horizons, with scatter points remaining tightly clustered around the ideal line even at *t*+43, and its regression line closely following the perfect reference. These analyses validate E2E-FANet's adaptability for short-, mid- and long-term forecasting. This stable performance indicates that the DBFM module can effectively extract frequency-domain features that are less affected by the prediction horizon, demonstrating its potential for deployment in challenging predictive scenarios.

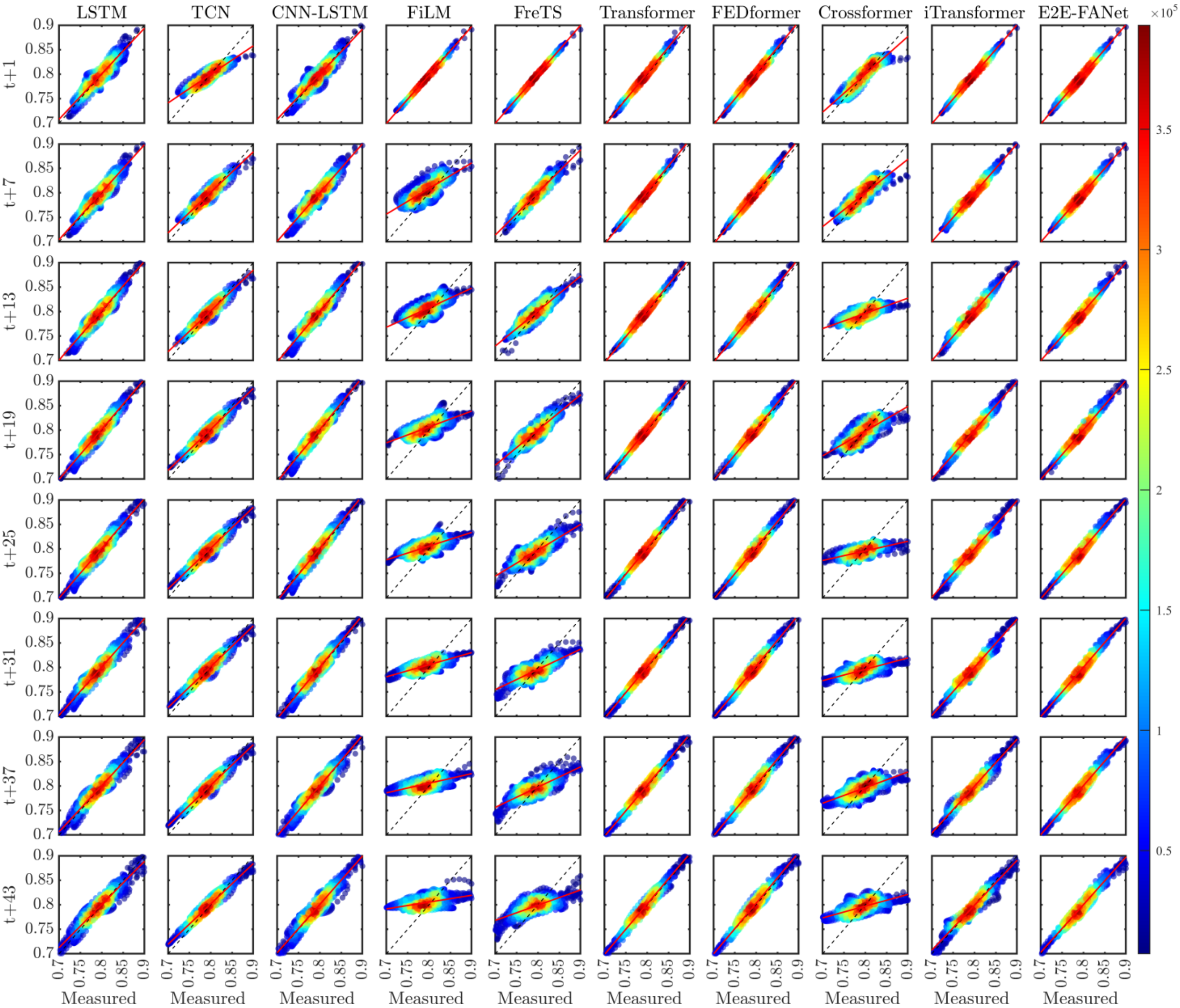

Fig. 7. Scatter density plots of E2E-FANet and baseline models at different prediction horizons.

### 3.5 Ablation experiment

To examine the individual contributions of each module within the E2E-FANet architecture and evaluate their synergistic effects, we conducted a systematic ablation study. We examined four distinct model configurations: a baseline model with only the exogenous-to-endogenous cross-attention module (E2ECA), a combination of temporal-wise attention and E2ECA modules (TA- E2ECA), an integration of dual-basis frequency mapping and E2ECA modules (DBFM- E2ECA), and the complete model incorporating all three modules (E2E-FANet).

Table 3 quantitatively compares the performance of these four configurations across MSE, MAE, RMSE, and MAPE metrics. The baseline E2ECA model achieved performance metrics of MSE = 0.0647, MAE = 0.1959, RMSE = 0.2543, and MAPE = 1.4260. As shown in Fig. 8(a), the scatter plot of E2ECA exhibits a dispersed distribution, indicating that predictions based solely on the E2ECA model contain considerable uncertainty. While E2ECA effectively captures the interactions between exogenous and endogenous variables, it lacks the capability to model the inherent temporal dynamics and frequency characteristics essential for accurate waves prediction. The incorporation of the TA module (TA-E2ECA) produced modest enhancements, reducing MSE by 0.46%, MAE by 1.89%, and RMSE by 0.16% compared to the baseline. Fig. 8(b) displays only marginally improved scatter plot concentration compared to Fig. 8(a), with the distribution remaining diffuse. This moderate improvement indicates that temporal attention, while helping the model focus on critical time steps, has limited effectiveness without explicit consideration of frequency-domain information. Notably, the DBFM-E2ECA configuration demonstrates substantial performance gains, reducing MSE by 39.13%, MAE by 26.79%, RMSE by 22.06%, and MAPE by 28.59% compared to TA-E2ECA. Fig. 8(c) reflects this advancement, showing a significantly more concentrated distribution along the diagonal line. The effectiveness of the DBFM module stems from its decomposition of wave signals into orthogonal cosine and sine basis functions, enabling the model to capture robust features related to wave periodicity. This structured frequency representation, combined with exogenous-to-endogenous cross-attention, allows DBFM-E2ECA to better represent the underlying wave physics than configurations without frequency awareness. The complete E2E-FANet achieved reductions of 12.00%, 3.34%, and 6.16% in MSE, MAE, and RMSE respectively compared to DBFM-E2ECA. While the MAPE showed a slight increase to 1.1113, the overall error metrics remained superior to all other configurations. Fig. 8(d) confirms this superior performance visually, showing the tightest clustering along the diagonal with minimal dispersion. These findings reveal three key insights. First, the DBFM module significantly contributes to error reduction. Second, the TA module effectively optimizes MSE. Third, the integration of all modules generates synergistic effects that markedly improve prediction accuracy. These results validate both the effectiveness and complementary nature of the proposed modules in wave forecasting applications.

Table 3 Ablation study results with different module configurations.

| Model | Module | | | Metric | | | |
|---|---|---|---|---|---|---|---|
| | DBFM | TA | E2ECA | MSE | MAE | RMSE | MAPE |
| E2ECA | | | √ | 0.0647 | 0.1959 | 0.2543 | 1.4260 |
| TA- E2ECA | | √ | √ | 0.0644 | 0.1922 | 0.2539 | 1.4252 |
| DBFM- E2ECA | √ | | √ | 0.0392 | 0.1407 | 0.1979 | **1.0176** |
| E2E-FANet | √ | √ | √ | **0.0345** | **0.1360** | **0.1857** | 1.1113 |

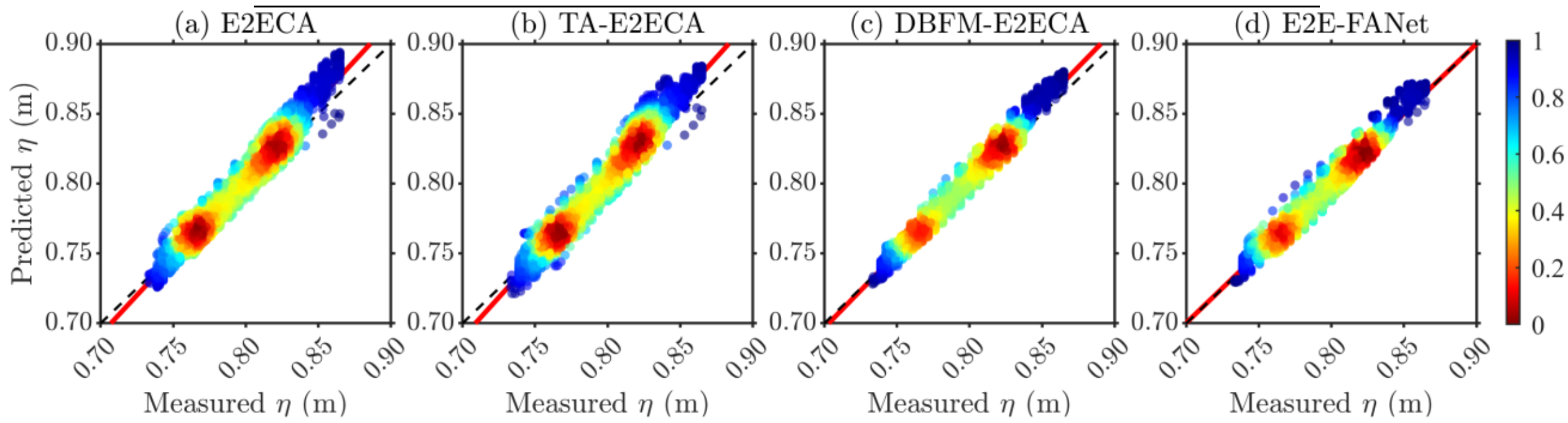

Fig. 8. Scatter density plots for ablation study configurations of E2E-FANet.

## 4. Generalization testing across different wave configurations

### 4.1 Evaluation of generalization under different wave conditions

To evaluate model generalization capabilities, we tested performance across seven datasets with varying $H_s$ and $T_p$ combinations: $H_s$ = 0.09 m, $T_p$ = 2 s; $H_s$ = 0.12 m, $T_p$ = 2 s; $H_s$ = 0.15 m, $T_p$ = 1.5 s; $H_s$ = 0.15 m, $T_p$ = 2 s; $H_s$ = 0.15 m, $T_p$ = 2.5 s; $H_s$ = 0.18 m, $T_p$ = 1.5 s; and $H_s$ = 0.21 m, $T_p$ = 2 s. These combinations were selected to encompass diverse ocean conditions. Using pre-trained weights from the main experiments, we generated 48-step predictions for each scenario, maintaining consistent evaluation metrics including MSE, MAE, RMSE, and MAPE. Table 4 presents model performance across all scenarios, with the best and second-best results in bold and underlined, respectively. Fig. 9(a) illustrates comparative performance through horizontally stacked bar charts, with each color representing a specific $H_s$ - $T_p$ combination. Analysis of Table 4 and Fig. 9(a) reveals E2E-FANet's superior performance across all scenarios. Quantitatively, E2E-FANet achieves average relative improvements of 27.91% in MSE, 18.68% in MAE, 15.21% in RMSE, and 20.37% in MAPE compared to the second-best model.

Table 4 Comparative performance metrics of models under various wave conditions.

| Datasets | Metric | LSTM | TCN | CNN -LSTM | FiLM | FreTS | Trans former | FED former | Cross former | iTrans former | E2E-FANet |
|---|---|---|---|---|---|---|---|---|---|---|---|
| | MSE | 0.0290 | 0.0194 | 0.0243 | 0.2150 | 0.1077 | 0.0038 | 0.0085 | 0.1832 | 0.0095 | **0.0028** |
| $H_s$ = 0.09 m | MAE | 0.1295 | 0.1093 | 0.1170 | 0.3647 | 0.2404 | 0.0494 | 0.0732 | 0.3349 | 0.0751 | **0.0402** |
| $T_p$ = 2 s | RMSE | 0.1703 | 0.1393 | 0.1558 | 0.4636 | 0.3283 | 0.0619 | 0.0924 | 0.4280 | 0.0979 | **0.0534** |
| | MAPE | 4.0504 | 2.4662 | 3.3500 | 7.8730 | 4.2976 | 1.7959 | 2.0684 | 6.3990 | 2.9126 | **0.9086** |
| | MSE | 0.0481 | 0.0298 | 0.0408 | 0.3700 | 0.1801 | 0.0065 | 0.0110 | 0.3077 | 0.0145 | **0.0041** |
| $H_s$ = 0.12 m | MAE | 0.1663 | 0.1346 | 0.1509 | 0.4797 | 0.3104 | 0.0639 | 0.0816 | 0.4325 | 0.0916 | **0.0487** |
| $T_p$ = 2 s | RMSE | 0.2193 | 0.1727 | 0.2020 | 0.6083 | 0.4244 | 0.0807 | 0.1052 | 0.5547 | 0.1205 | **0.0644** |
| | MAPE | 1.9244 | 1.2170 | 2.0786 | 4.2985 | 3.4881 | 0.7510 | 0.9750 | 4.2401 | 1.0187 | **0.5922** |
| | MSE | 0.1344 | 0.0594 | 0.1056 | 0.6199 | 0.3804 | 0.0129 | 0.0180 | 0.5680 | 0.0570 | **0.0109** |
| $H_s$ = 0.15 m | MAE | 0.2723 | 0.1821 | 0.2363 | 0.6021 | 0.4411 | 0.0860 | 0.1033 | 0.5716 | 0.1636 | **0.0755** |
| $T_p$ = 1.5 s | RMSE | 0.3666 | 0.2437 | 0.3249 | 0.7873 | 0.6167 | 0.1136 | 0.1344 | 0.7536 | 0.2387 | **0.1044** |
| | MAPE | 2.4000 | 1.3465 | 2.0494 | 3.3827 | 2.8006 | 0.7769 | 1.0477 | 2.945 | 1.2650 | **0.7483** |
| | MSE | 0.0690 | 0.0458 | 0.0611 | 0.5436 | 0.2611 | 0.0104 | 0.0171 | 0.4549 | 0.0215 | **0.0059** |
| $H_s$ = 0.15 m | MAE | 0.1993 | 0.1656 | 0.1859 | 0.5817 | 0.3742 | 0.0799 | 0.1008 | 0.5252 | 0.1115 | **0.0577** |
| $T_p$ = 2 s | RMSE | 0.2627 | 0.2141 | 0.2473 | 0.7373 | 0.5109 | 0.1022 | 0.1306 | 0.6744 | 0.1466 | **0.0772** |
| | MAPE | 1.7622 | 1.4165 | 1.7184 | 3.4935 | 2.5674 | 0.6739 | 0.9747 | 2.9938 | 0.9774 | **0.5186** |
| | MSE | 0.1106 | 0.1268 | 0.1106 | 0.9195 | 0.5570 | 0.0347 | 0.0670 | 0.7299 | 0.0957 | **0.0302** |
| $H_s$ = 0.15 m | MAE | 0.2392 | 0.2618 | 0.2366 | 0.7366 | 0.5455 | 0.1333 | 0.1922 | 0.6556 | 0.2166 | **0.1200** |
| $T_p$ = 2.5 s | RMSE | 0.3325 | 0.3562 | 0.3326 | 0.958 | 0.7463 | 0.1864 | 0.2589 | 0.8543 | 0.3094 | **0.1738** |
| | MAPE | 4.0504 | 2.4662 | 3.3500 | 7.8730 | 4.2976 | 1.7959 | 2.0684 | 6.3990 | 2.9126 | **0.9086** |
| | MSE | 0.1780 | 0.0837 | 0.1382 | 0.8770 | 0.5182 | 0.0231 | 0.0270 | 0.8005 | 0.0808 | **0.0184** |
| $H_s$ = 0.18 m | MAE | 0.3183 | 0.2165 | 0.2739 | 0.7334 | 0.5223 | 0.1150 | 0.1254 | 0.6953 | 0.1988 | **0.0990** |
| $T_p$ = 1.5 s | RMSE | 0.4219 | 0.2894 | 0.3718 | 0.9364 | 0.7199 | 0.1520 | 0.1644 | 0.8947 | 0.2843 | **0.1359** |
| | MAPE | 4.6975 | 1.5534 | 3.8470 | 4.0714 | 4.1462 | 2.2317 | 1.9286 | 4.2916 | **1.4696** | 1.5688 |
| | MSE | 0.1340 | 0.1075 | 0.1278 | 0.9848 | 0.5065 | 0.0387 | 0.0535 | 0.8561 | 0.0557 | **0.0232** |
| $H_s$ = 0.21 m | MAE | 0.2757 | 0.2432 | 0.2688 | 0.7825 | 0.5235 | 0.1398 | 0.1699 | 0.7199 | 0.1722 | **0.1056** |
| $T_p$ = 2 s | RMSE | 0.3661 | 0.3278 | 0.3574 | 0.9924 | 0.7117 | 0.1969 | 0.2314 | 0.9252 | 0.2361 | **0.1523** |
| | MAPE | 1.7732 | 2.0482 | 1.9571 | 3.8414 | 3.0197 | 0.9015 | 1.5257 | 3.1955 | 1.1531 | **0.8778** |

Wave parameters $H_s$ significantly influence prediction performance, with larger values typically indicating more complex ocean conditions and stronger nonlinear characteristics. Analysis of performance improvements across different $H_s$ conditions reveals a distinctive pattern in E2E-FANet's generalization performance with varying $H_s$. While the E2E-FANet model consistently outperforms baselines across all $H_s$ conditions, the magnitude of these improvements varies systematically with $H_s$. Under low $H_s$ conditions with $H_s$ at 0.09 m, E2E-FANet achieves its most substantial performance gains, with average relative improvements of 31.62% in MSE, 21.21% in MAE, 16.97% in RMSE, and 35.28% in MAPE compared to the second-best model. These results indicate that the E2E-FANet exhibits superior performance under low $H_s$ conditions. For medium $H_s$ conditions with $H_s$ at 0.12 m and $H_s$ at 0.15 m, E2E-FANet maintains clear superiority, though with reduced margins of improvement: 23.91% in MSE, 16.66% in MAE, 13.11% in RMSE, and 25.38% in MAPE. Notably, under high $H_s$ conditions with $H_s$ at 0.18 m and $H_s$ at 0.21 m, E2E-FANet's performance advantages rebound, particularly in MSE and RMSE metrics, approaching levels observed in low $H_s$ scenarios, with specific improvements of 30.20% in MSE, 19.19% in MAE, 16.62% in RMSE, and 16.17% in MAPE. These results indicate that E2E-FANet performs exceptionally well across varying $H_s$ conditions. Consequently, E2E-FANet demonstrates strong generalization capability across diverse wave scenarios.

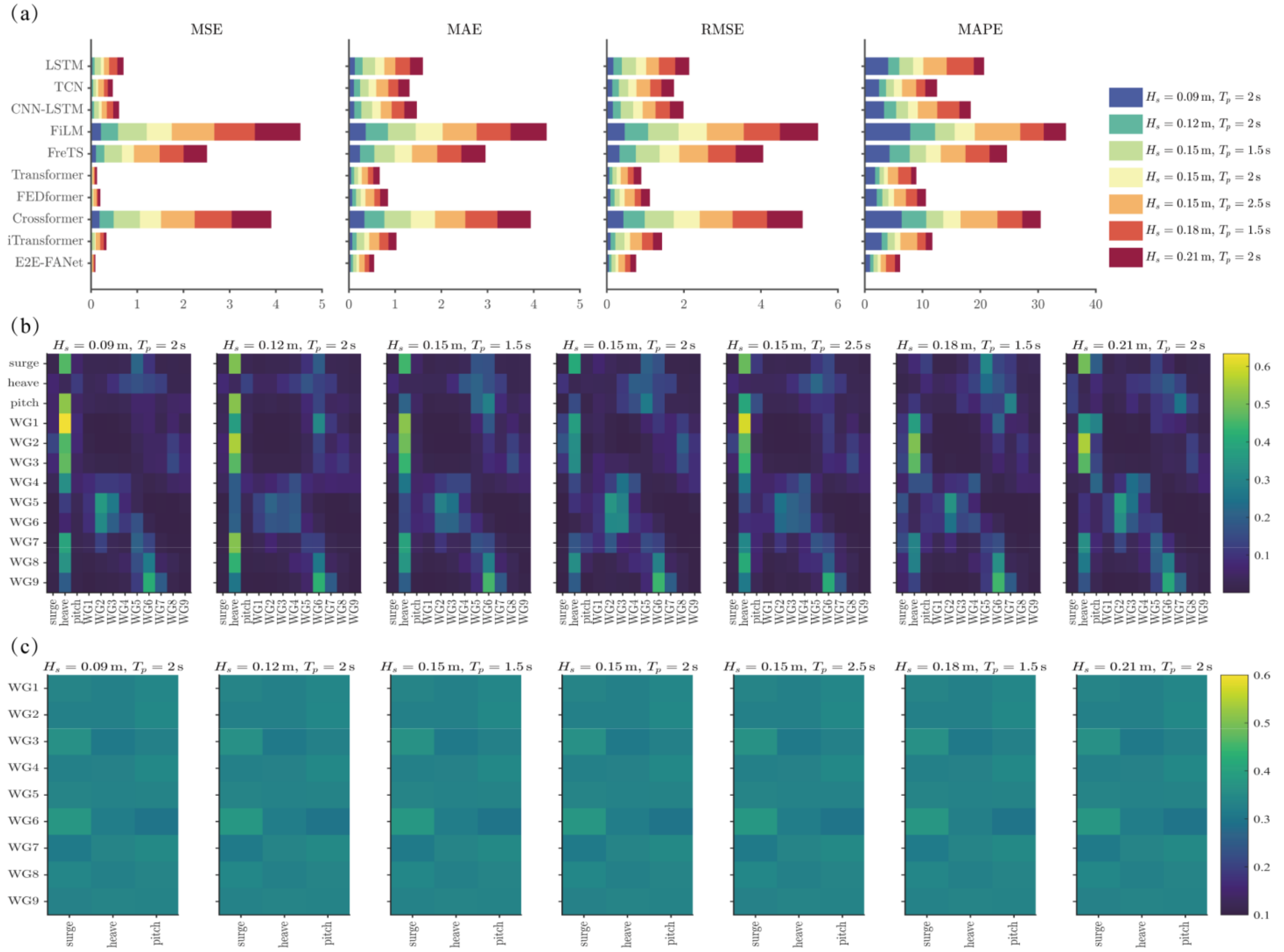

Fig. 9. Analysis of model generalization and attention mechanisms in different novel scenarios.

The robust performance of E2E-FANet across varying $H_s$ conditions stems from its specialized architecture, particularly the E2ECA module. To elucidate the mechanisms underlying this robust generalization, we analyze the attention-weight visualization heatmaps in Fig. 9(b) and 9(c), comparing iTransformer's variable self-attention distribution with E2ECA's variate-wise cross-attention distribution. The heatmaps plot input variables on both axes, where WG1-WG9 represent wave gauges, which are endogenous variables. Surge, heave, and pitch represent the motion responses of the floating breakwater, which are exogenous variables. Fig. 9(b) shows that iTransformer's attention patterns focus on a few variables under each wave condition, possibly because of their stronger signals. This creates distinct attention blocks and ignores other variables. For instance, in the $H_s = 0.12$ m, $T_p = 2$ s scenario, iTransformer primarily attends to WG1 and WG2, while ignoring other input variables. In contrast, E2ECA, as shown in Fig. 9(c), allocates attention across all input variables under varying wave conditions. This balanced attention strategy enables E2E-FANet to effectively utilize information from all input variables, regardless of wave conditions, resulting in more robust and generalizable predictions. Experimental results across seven novel scenarios validate E2E-FANet's exceptional generalization capability.

## 4.2 Ablation study on model components for generalization capability

To investigate the contribution of each module in the E2E-FANet architecture to the model's generalization ability, we conducted comparative analyses across seven datasets with different combinations of $H_s$ and $T_p$, maintaining consistency with our main experimental conditions. Table 5 presents comparative results, where bold and underlined values denote the best and second-best performances, respectively, while Fig. 10 provides visual representations of these comparisons. The height of each 3D bar represents the magnitude of the corresponding performance metric. The baseline E2ECA model demonstrates a certain degree of generalization capability, but its performance is limited. This indicates that although cross-attention alone can capture interactions between variables, it is insufficient for achieving robust generalization. The DBFM-E2ECA model demonstrates the most significant improvement in generalization capability. Fig. 10 shows that the bar graphs of DBFM-E2ECA decrease noticeably across all subplots compared to previous configurations. This indicates that the DBFM component is a key driver of E2E-FANet's generalization ability, enabling the model to capture underlying wave patterns independent of specific wave parameters. The complete E2E-FANet model demonstrates superior performance metric values across all indicators and wave conditions. This advantage is particularly evident in the MSE and RMSE metrics. For example, Table 5 shows that under wave conditions of $H_s$ = 0.09 m and $T_p$ = 2 s, E2ECA achieves an MSE of 0.0152, while E2E-FANet significantly reduces this to 0.0028, representing an 81.57% reduction in error. In the MSE subplot of Fig.10, the notably shorter bar for E2E-FANet visually illustrates this substantial reduction. This comprehensive analysis, combining graphical evidence and numerical validation, clearly demonstrates the specific contribution of each module to the generalization capability of the E2E-FANet framework. The results validate the effectiveness of our proposed DBFM and E2ECA modules in enhancing generalization performance.

Table 5 Comparative performance metrics with different module configurations.

| Datasets | Metric | Model | | | |
|---|---|---|---|---|---|
| | | E2ECA | TA- E2ECA | DBFM- E2ECA | E2E-FANet |
| $H_s$ = 0.09 m | MSE | 0.0152 | 0.0159 | <u>0.0045</u> | **0.0028** |
| $T_p$ = 2 s | MAE | 0.0966 | 0.0929 | <u>0.0505</u> | **0.0402** |
| | RMSE | 0.1234 | 0.1263 | <u>0.0669</u> | **0.0534** |
| | MAPE | 2.7557 | 2.7846 | <u>1.5722</u> | **0.9086** |
| $H_s$ = 0.12 m | MSE | 0.0183 | 0.0170 | <u>0.0056</u> | **0.0041** |
| $T_p$ = 2 s | MAE | 0.1055 | 0.0970 | <u>0.0567</u> | **0.0487** |
| | RMSE | 0.1351 | 0.1303 | <u>0.0749</u> | **0.0644** |
| | MAPE | 1.5238 | 1.3116 | <u>0.6941</u> | **0.5922** |
| $H_s$ = 0.15 m | MSE | 0.0389 | 0.0400 | <u>0.0119</u> | **0.0109** |
| $T_p$ = 1.5 s | MAE | 0.1480 | 0.1509 | <u>0.0789</u> | **0.0755** |
| | RMSE | 0.1974 | 0.2000 | <u>0.1090</u> | **0.1044** |
| | MAPE | 1.4044 | 1.5305 | <u>0.7756</u> | **0.7483** |
| $H_s$ = 0.15 m | MSE | 0.0232 | 0.0198 | <u>0.0078</u> | **0.0059** |
| $T_p$ = 2 s | MAE | 0.1189 | 0.1066 | <u>0.0664</u> | **0.0577** |
| | RMSE | 0.1523 | 0.1408 | <u>0.0883</u> | **0.0772** |
| | MAPE | 1.1061 | 0.9753 | <u>0.6411</u> | **0.5186** |
| $H_s$ = 0.15 m | MSE | 0.1012 | 0.0945 | <u>0.0336</u> | **0.0302** |
| $T_p$ = 2.5 s | MAE | 0.2393 | 0.2281 | <u>0.1293</u> | **0.1200** |
| | RMSE | 0.3182 | 0.3074 | 0.1834 | **0.1738** |
| | MAPE | 1.7323 | 1.6653 | **1.1930** | 1.2385 |
| $H_s$ = 0.18 m | MSE | 0.0537 | 0.0575 | **0.0176** | 0.0184 |
| $T_p$ = 1.5 s | MAE | 0.1738 | 0.1831 | **0.0972** | 0.0990 |
| | RMSE | 0.2317 | 0.2399 | **0.1326** | 0.1359 |
| | MAPE | 2.6307 | 2.4191 | 1.5814 | **1.5688** |
| $H_s$ = 0.21 m | MSE | 0.0563 | 0.0538 | 0.0243 | **0.0232** |
| $T_p$ = 2 s | MAE | 0.1773 | 0.1731 | 0.1085 | **0.1056** |
| | RMSE | 0.2372 | 0.2319 | 0.1560 | **0.1523** |
| | MAPE | 1.3610 | 1.3690 | 0.8961 | **0.8778** |

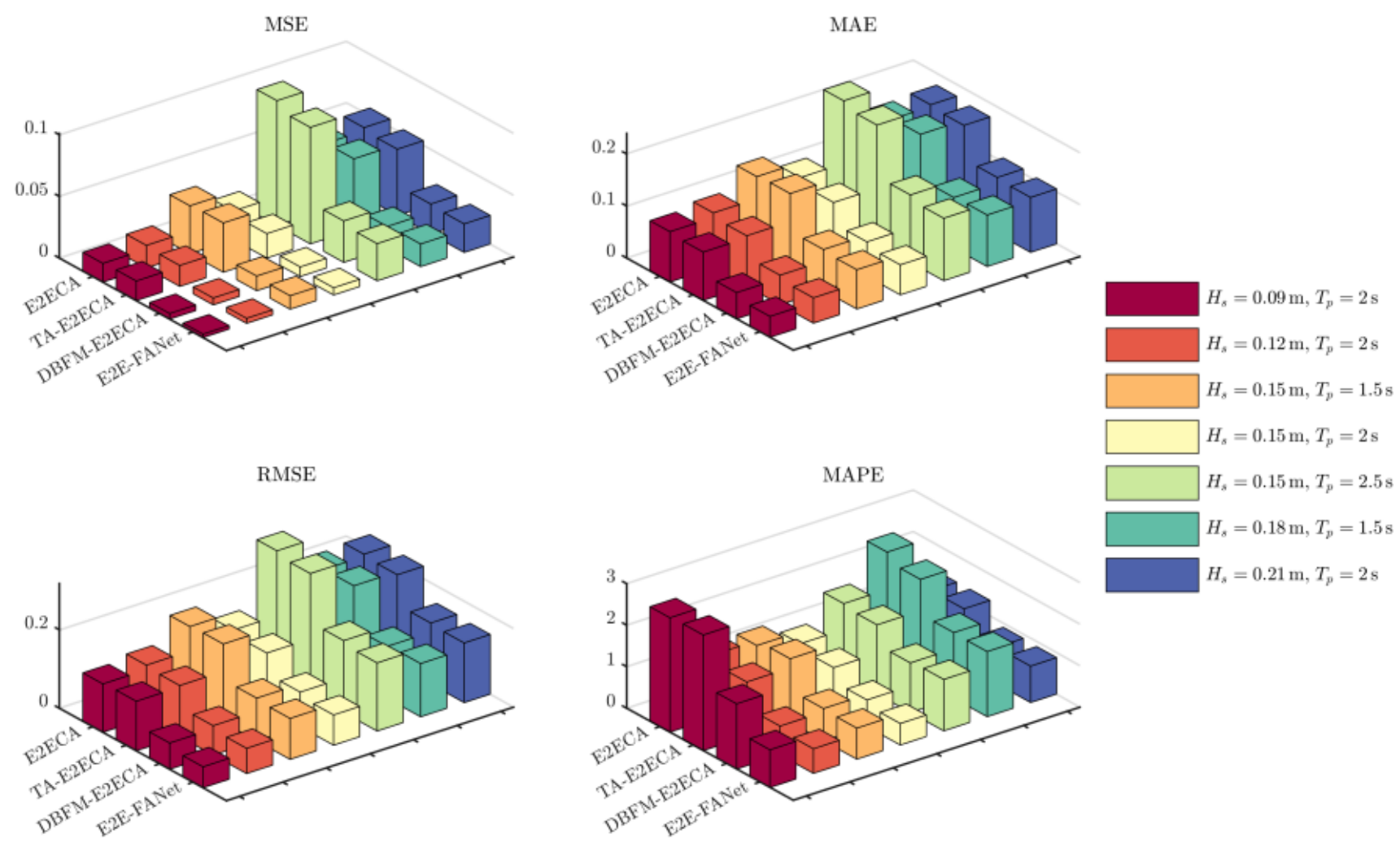

Fig. 10. Performance metrics for ablation configurations across wave conditions.

## 4.3 Hyperparameter optimization for model performance and generalization

To optimize the E2E-FANet's performance and ensure robust generalization across diverse datasets, we conducted systematic hyperparameter experiments. The investigation focused on two key aspects: the influence of exogenous variable number, and the optimal layer number. The experimental framework consisted of two phases: first, evaluating baseline performance under $H_s$ = 0.18 m, $T_p$ = 2 s conditions, followed by assessing generalization capabilities across different conditions.

### 4.3.1 Influence of the number of exogenous variables

We investigated the impact of exogenous variable quantity on prediction accuracy by evaluating three model variants, each incorporating a different number of exogenous variables under the conditions of $H_s$ = 0.18 m, $T_p$ = 2 s conditions. The endogenous variables remained constant, while we incrementally added exogenous variables. We evaluated model performance using four metrics: MSE, MAE, RMSE, and MAPE. Fig. 11 visually presents the impact of varying the number of exogenous variables on E2E-FANet's prediction performance. As shown in Fig. 11, the bar heights decrease when increasing from one to two exogenous variables, suggesting improved performance with two variables. However, further increase to three variables shows a slight increase in bar heights across most metrics. To provide comprehensive insight into the influence of exogenous variable quantity, Fig. 12 presents scatter density plots comparing predicted and measured values across model variants. In each scatter plot of Fig. 12, the horizontal axis represents measured values, the vertical axis shows model predictions, and a dashed black line indicates the ideal prediction line (y=x). The density of data points is represented by a color gradient from blue to red. The scatter density plot for the three-variable configuration demonstrates the most concentrated point clustering around the ideal diagonal line, indicating higher prediction accuracy compared to one- and two-variable models. While the two-variable model shows strong clustering, the three-variable model achieves a more concentrated distribution along the diagonal, particularly in high-density regions.

To resolve this apparent discrepancy and evaluate generalization ability, we conducted comprehensive testing across diverse datasets, with results documented in Table 6. The three-variable model consistently outperformed other configurations across multiple wave conditions. The advantages were particularly pronounced when predicting challenging scenarios such as $H_s$ = 0.21 m, $T_p$ = 2 s, or $H_s$ = 0.15 m, $T_p$ = 2.5 s. Through this comprehensive analysis, we observed that while the introduction of a third exogenous variable did not lead to improvements under standard conditions, it significantly enhanced prediction accuracy across varying fluctuation conditions. Based on this enhanced adaptability to diverse wave conditions, we adopted the three-variable configuration as the optimal architecture for our model implementation.

Table 6 Comparative performance under different wave conditions for models using one, two, and three exogenous variables.

| Datasets | Metric | Number of exogenous variables | | |
|---|---|---|---|---|
| | | 1 | 2 | 3 |
| $H_s$ = 0.09 m | MSE | 0.0032 | 0.0032 | **0.0028** |
| $T_p$ = 2 s | MAE | 0.0434 | 0.0432 | **0.0402** |
| | RMSE | 0.0565 | 0.0564 | **0.0534** |
| | MAPE | 1.2275 | 0.9760 | **0.9086** |
| $H_s$ = 0.12 m | MSE | 0.0043 | 0.0045 | **0.0041** |
| $T_p$ = 2 s | MAE | 0.0503 | 0.0512 | **0.0487** |
| | RMSE | 0.0660 | 0.0671 | **0.0644** |
| | MAPE | **0.5884** | 0.6267 | 0.5922 |
| $H_s$ = 0.15 m | MSE | **0.0109** | 0.0115 | **0.0109** |
| $T_p$ = 1.5 s | MAE | 0.0764 | 0.0780 | **0.0755** |
| | RMSE | **0.1042** | 0.1070 | 0.1044 |
| | MAPE | 0.7924 | 0.7584 | **0.7483** |
| $H_s$ = 0.15 m | MSE | 0.0063 | 0.0064 | **0.0059** |
| $T_p$ = 2 s | MAE | 0.0602 | 0.0604 | **0.0577** |
| | RMSE | 0.0794 | 0.0801 | **0.0772** |
| | MAPE | 0.5796 | 0.5509 | **0.5186** |
| $H_s$ = 0.15 m | MSE | 0.0312 | 0.0321 | **0.0302** |
| $T_p$ = 2.5 s | MAE | 0.1230 | 0.1249 | **0.1200** |
| | RMSE | 0.1766 | 0.1791 | **0.1738** |
| | MAPE | **1.1901** | 1.2536 | 1.2385 |
| $H_s$ = 0.18 m | MSE | **0.0183** | 0.0191 | 0.0184 |
| $T_p$ = 1.5 s | MAE | **0.0990** | 0.1015 | **0.0990** |
| | RMSE | **0.1353** | 0.1382 | 0.1359 |
| | MAPE | 1.5912 | **1.4597** | 1.5688 |
| $H_s$ = 0.21 m | MSE | 0.0244 | 0.0244 | **0.0232** |
| $T_p$ = 2 s | MAE | 0.1102 | 0.1096 | **0.1056** |
| | RMSE | 0.1562 | 0.1562 | **0.1523** |
| | MAPE | 0.9203 | 0.9057 | **0.8778** |

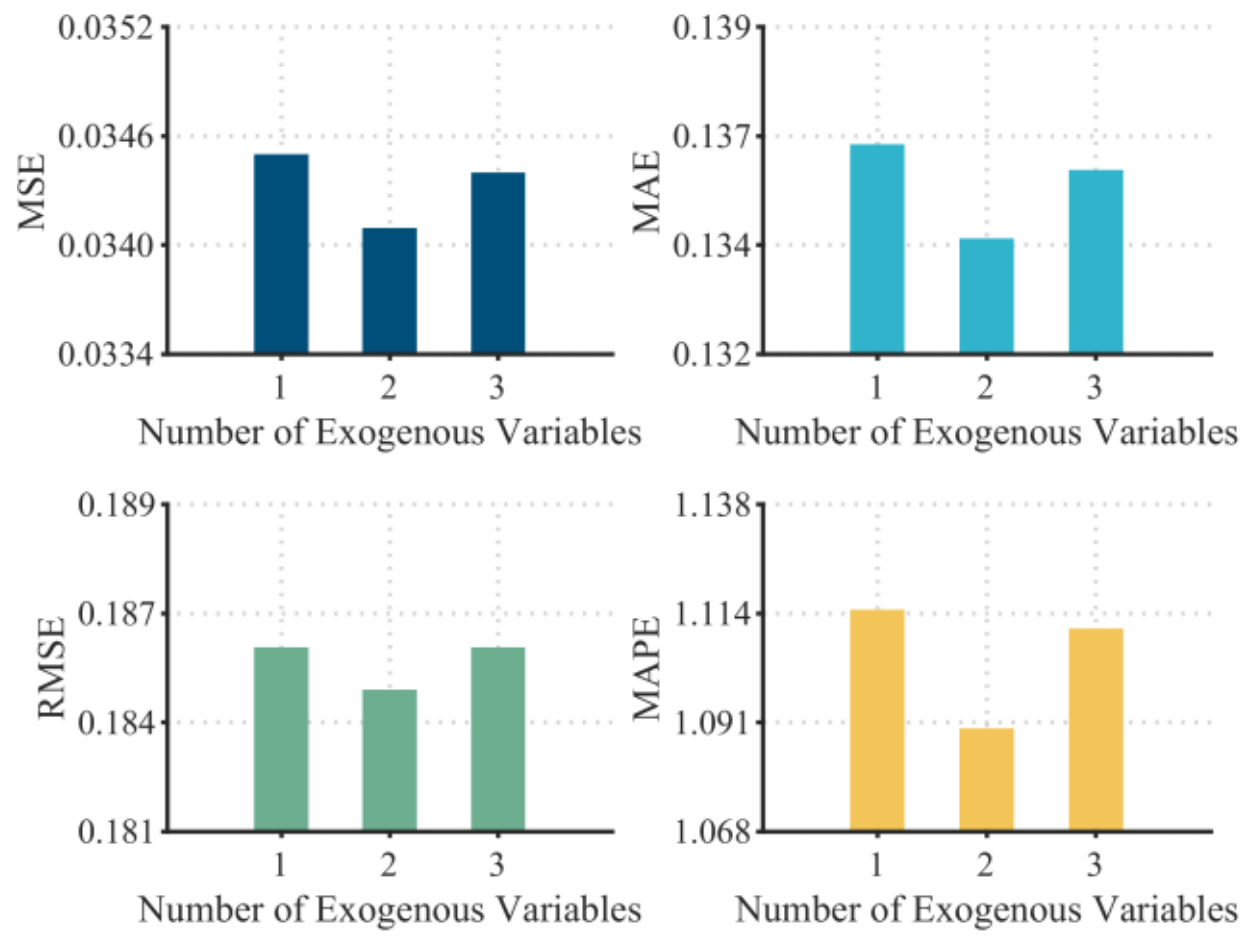

Fig. 11. Bar chart comparison of performance metrics with varying number of exogenous variables.

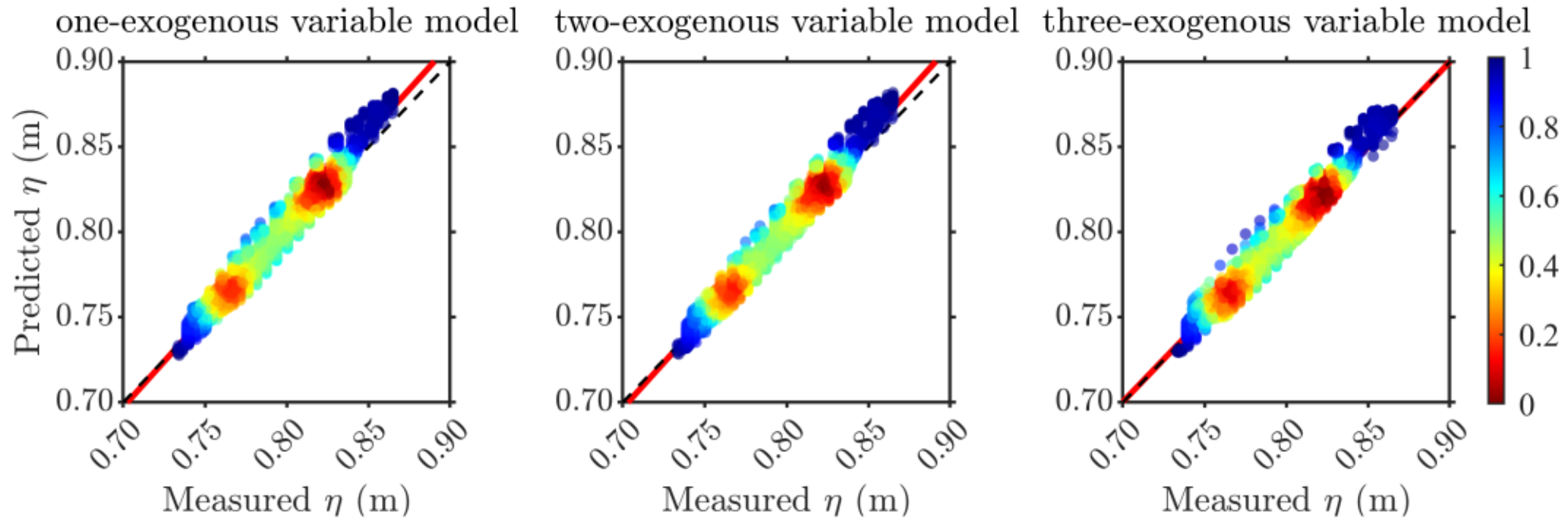

Fig. 12. Scatter density plots for models with varying numbers of exogenous variables.

#### 4.3.2 Influence of the number of layers

We examined the influence of layer depth by evaluating architectures with one to four layers. We evaluated model performance using four metrics: MSE, MAE, RMSE, and MAPE. Fig. 13 visually illustrates the impact of varying the number of layers on E2E-FANet's prediction performance. The analysis demonstrates a significant performance improvement when increasing from one to two layers. However, the addition of a third layer yields only marginal improvements in select metrics, suggesting limited additional learning benefits while increasing computational burden. Further expansion to four layers leads to performance degradation, due to overfitting or vanishing gradient problems. The scatter density plots in Fig. 14 provide complementary insight by illustrating the comparison between predicted and measured values across different network depths. The distribution patterns show progressive convergence toward the diagonal line as depth increases from one to two layers, with the two-layer configuration exhibiting optimal point density concentration. However, the four-layer configuration displays significant deviation from the diagonal line with a more scattered density distribution, confirming the performance decline observed in the metric analysis.

To validate the generalization ability of models with varying layer depth, we conducted comprehensive testing across diverse datasets, with results documented in Table 7. The two-layer configuration consistently outperformed other architectures across various wave scenarios. Notable improvements were particularly evident in datasets with $H_s$ = 0.12 m, $T_p$ = 2 s, and $H_s$ = 0.09 m, $T_p$ = 2 s. This consistent performance across diverse conditions indicates that the two-layer architecture effectively balances feature extraction capability with computational efficiency, avoiding the potential instability associated with deeper networks while maintaining robust predictive capability. Based on this comprehensive evaluation, we established the two-layer architecture as the optimal network depth configuration for our model implementation.

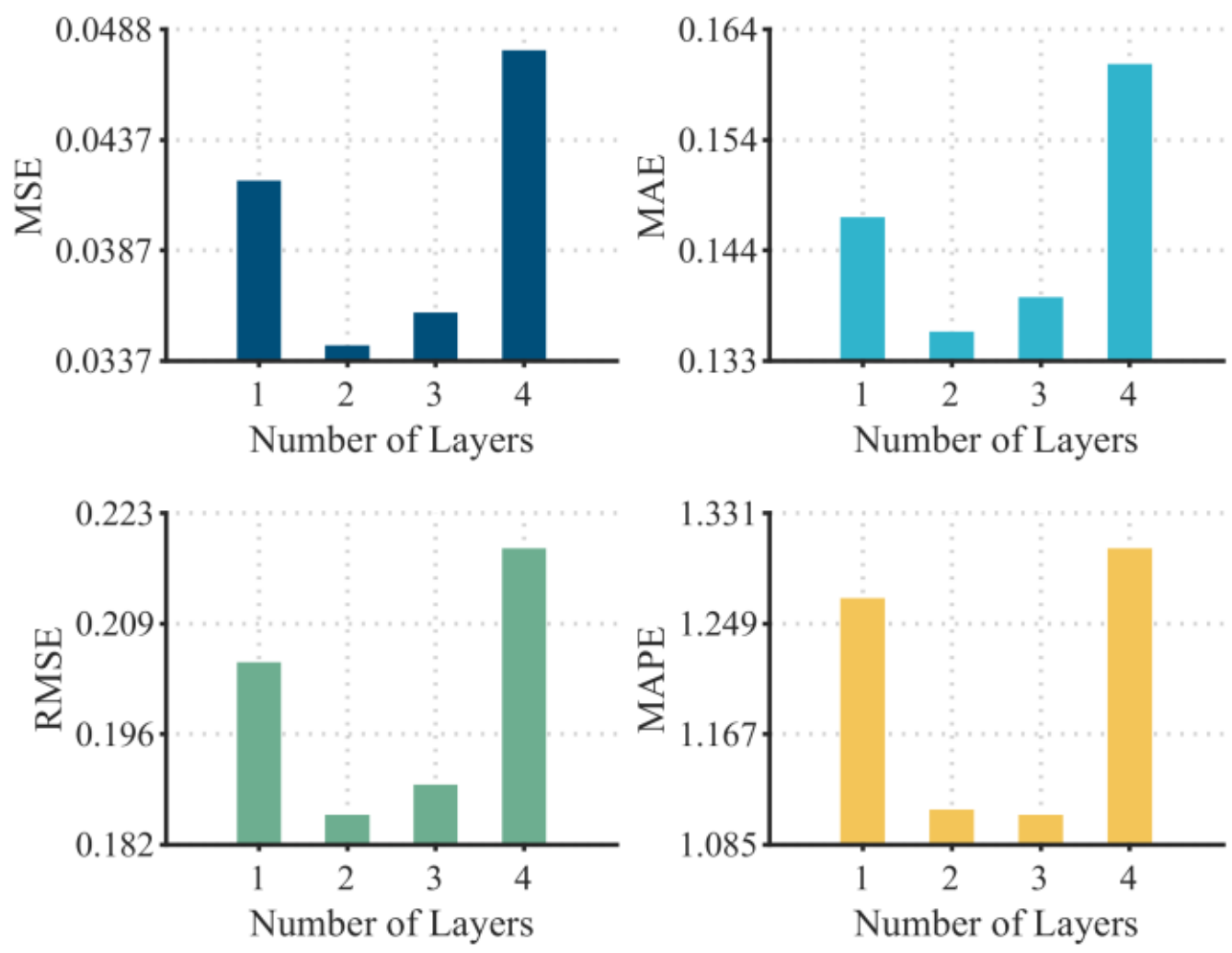

Fig. 13. Bar chart comparison of performance metrics with varying layer depth.

Table 7 Comparative performance under different wave conditions for models using one, two, three, and four layers.

| Datasets | Metric | Layer | | | |
|---|---|---|---|---|---|
| | | 1 | 2 | 3 | 4 |
| $H_s$ = 0.09 m | MSE | 0.0045 | **0.0028** | 0.0037 | 0.0053 |
| $T_p$ = 2 s | MAE | 0.0522 | **0.0402** | 0.0472 | 0.0566 |
| | RMSE | 0.0668 | **0.0534** | 0.0609 | 0.0727 |
| | MAPE | 1.3639 | **0.9086** | 1.1422 | 1.2914 |
| $H_s$ = 0.12 m | MSE | 0.0061 | **0.0041** | 0.0052 | 0.0074 |
| $T_p$ = 2 s | MAE | 0.0602 | **0.0487** | 0.0553 | 0.0667 |
| | RMSE | 0.0780 | **0.0644** | 0.0718 | 0.0862 |
| | MAPE | 0.7765 | **0.5922** | 0.6726 | 0.8865 |
| $H_s$ = 0.15 m | MSE | 0.0140 | **0.0109** | 0.0121 | 0.0172 |
| $T_p$ = 1.5 s | MAE | 0.0864 | **0.0755** | 0.0809 | 0.0973 |
| | RMSE | 0.1182 | **0.1044** | 0.1100 | 0.1311 |
| | MAPE | 0.8024 | **0.7483** | 0.8415 | 0.9456 |
| $H_s$ = 0.15 m | MSE | 0.0087 | **0.0059** | 0.0072 | 0.0106 |
| $T_p$ = 2 s | MAE | 0.0715 | **0.0577** | 0.0647 | 0.0787 |
| | RMSE | 0.0935 | **0.0772** | 0.0851 | 0.1028 |
| | MAPE | 0.6734 | **0.5186** | 0.5913 | 0.7196 |
| $H_s$ = 0.15 m | MSE | 0.0410 | **0.0302** | 0.0324 | 0.0419 |
| $T_p$ = 2.5 s | MAE | 0.1421 | **0.1200** | 0.1273 | 0.1467 |
| | RMSE | 0.2025 | **0.1738** | 0.1799 | 0.2048 |
| | MAPE | **1.1787** | 1.2385 | 1.2944 | 1.3550 |
| $H_s$ = 0.18 m | MSE | 0.0227 | **0.0184** | 0.0196 | 0.0271 |
| $T_p$ = 1.5 s | MAE | 0.1095 | **0.0990** | 0.1034 | 0.1222 |
| | RMSE | 0.1507 | **0.1359** | 0.1401 | 0.1647 |
| | MAPE | 1.9999 | **1.5688** | 1.7935 | 1.7579 |
| $H_s$ = 0.21 m | MSE | 0.0311 | **0.0232** | 0.0262 | 0.0351 |
| $T_p$ = 2 s | MAE | 0.1254 | **0.1056** | 0.1148 | 0.1348 |
| | RMSE | 0.1764 | **0.1523** | 0.1618 | 0.1874 |
| | MAPE | 1.0282 | **0.8778** | 0.9314 | 1.0538 |

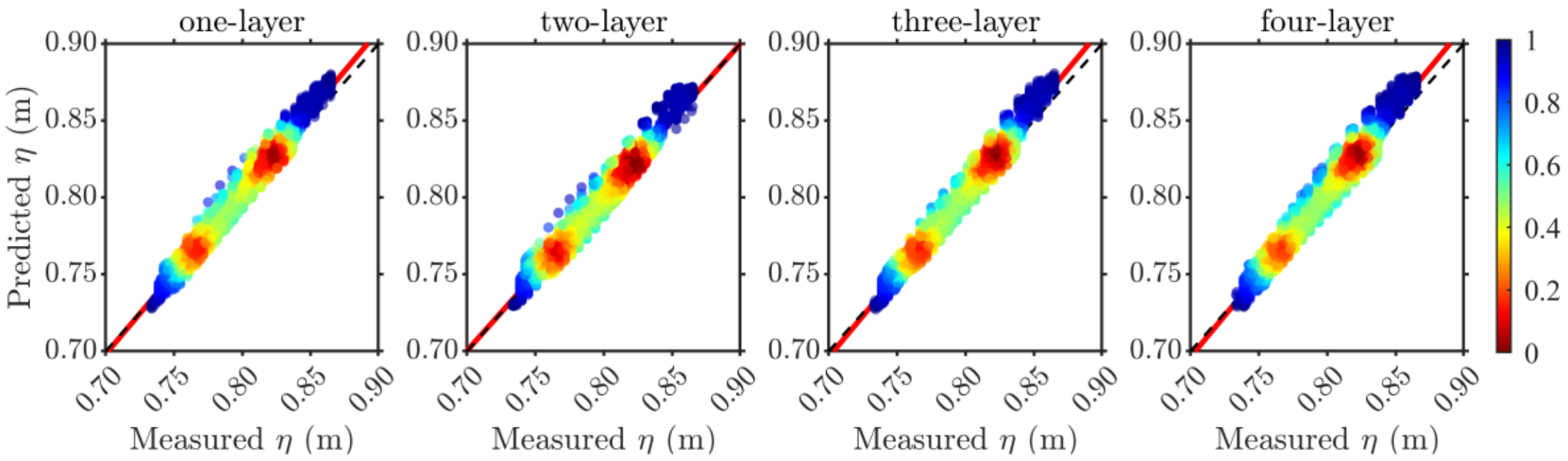

Fig. 14. Scatter density plots for models with varying layer depth.

## 5. Adaptability testing with varying relative water density

### 5.1 Fine-tuning results and analysis

To evaluate E2E-FANet's adaptability across varying FB, we implemented fine-tuning under different relative water density ($RW$) configurations. $RW$ is defined as the ratio of draft to water depth. The experimental framework focused on two representative scenarios: $RW = 0.8$, $H_s = 0.18$ m, $T_p = 2$ s, and $RW = 0.4$, $H_s = 0.18$ m, $T_p = 2$ s. The fine-tuning process utilized pre-trained models from main experiments, maintaining consistent data partitioning ratios and evaluation metrics to ensure comparative validity. The fine-tuning strategy specifically targeted parameter optimization in the final linear layer while preserving pre-trained weights in preceding layers. To ensure a fair methodological comparison, we conducted analyses using only Transformer-based models as benchmarks, since E2E-FANet is also built on the Transformer architecture. This experiment was designed to evaluate model adaptability across diverse $RW$ conditions, quantify transfer learning effectiveness through fine-tuning approaches, and benchmark performance against other established Transformer architectures for a fair comparison within the same architectural family.

As shown in Table 8, we evaluated five models for waves prediction under two distinct $RW$ conditions, utilizing four performance metrics: MSE, MAE, RMSE, and MAPE. The comprehensive results demonstrate E2E-FANet's superior performance across both RW conditions. Under $RW = 0.8$, E2E-FANet achieved exceptional results with MSE of 0.0188, MAE of 0.1041, RMSE of 0.1372, and MAPE of 0.6017, showing significant improvements over the second-best performing model, Transformer, with reductions of 36.70% in MSE, 21.85% in MAE, 20.33% in RMSE, and 27.10% in MAPE. Similarly, for $RW = 0.4$ conditions, E2E-FANet maintained its performance advantage, outperforming Transformer with reductions of 34.68%, 19.65%, 19.18%, and 8.86% across respective metrics. The results demonstrate that E2E-FANet maintains high accuracy in waves predictions across $RW$ conditions, thereby validating the model's adaptability to diverse $RW$ conditions. This reflects E2E-FANet can capture the essential characteristics of floating structure dynamics, demonstrating significant potential for applications in dynamic scenarios requiring rapid environmental adaptation, such as emergency breakwater deployment.

Table 8 Fine-tuning performance on two $RW$ conditions.

| Datasets | Metric | Transformer | FEDformer | Crossformer | iTransformer | E2E-FANet |
|---|---|---|---|---|---|---|
| $RW = 0.8$, $H_s = 0.18$ m, $T_p = 2$ s | MSE | <u>0.0297</u> | 0.0407 | 0.5409 | 0.0452 | **0.0188** |
| | MAE | <u>0.1332</u> | 0.1595 | 0.5579 | 0.1590 | **0.1041** |
| | RMSE | <u>0.1722</u> | 0.2018 | 0.7355 | 0.2127 | **0.1372** |
| | MAPE | <u>0.8252</u> | 1.1234 | 2.1182 | 0.8840 | **0.6017** |
| $RW = 0.4$, $H_s = 0.18$ m, $T_p = 2$ s | MSE | <u>0.0346</u> | 0.0448 | 0.7132 | 0.0517 | **0.0226** |
| | MAE | <u>0.1410</u> | 0.1624 | 0.6354 | 0.1697 | **0.1133** |
| | RMSE | <u>0.1861</u> | 0.2117 | 0.8445 | 0.2275 | **0.1504** |
| | MAPE | <u>0.7570</u> | 1.0062 | 2.3071 | 0.8818 | **0.6899** |

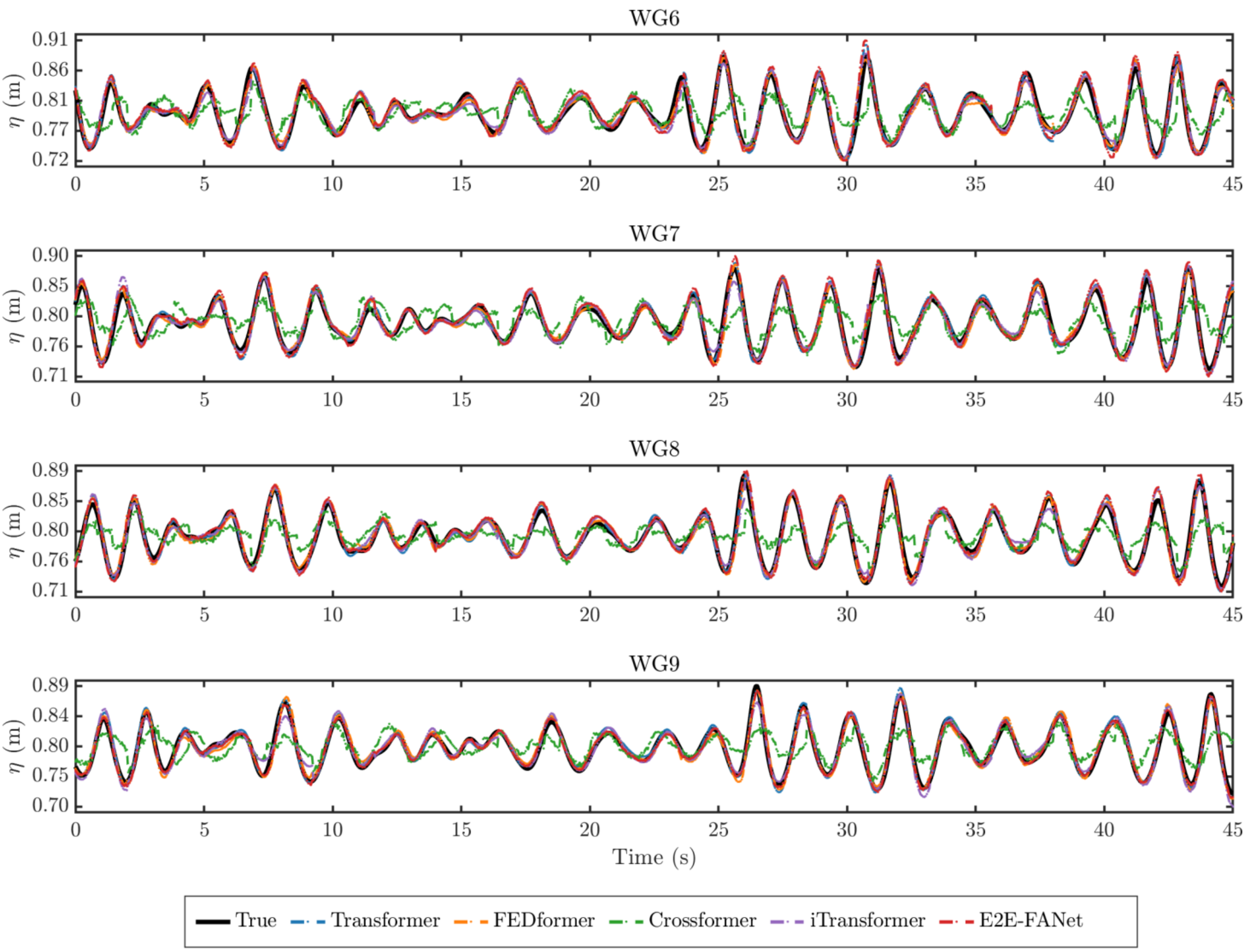

Fig. 15. Comparison of wave elevation predictions for *RW* = 0.8 condition at downstream wave gauges.

Fig. 15 and 16 present time-series comparisons of wave elevation predictions at varying *RW* conditions. The plots compare predictions from E2E-FANet and benchmark models at four downstream wave gauges (WG6-WG9) against measured wave elevations. The E2E-FANet prediction curves show the strongest agreement with measured wave elevation trajectories. While Transformer and FEDformer models adequately capture overall wave elevation patterns, they show notable deviations at wave extrema and during specific time periods. Crossformer and iTransformer exhibit larger discrepancies, with less consistent wave signal tracking and greater deviations across wave gauges. The results indicate that fine-tuning the pre-trained E2E-FANet enables effective adaptation to both low and high *RW* scenarios. These visual findings align with the quantitative metrics in Table 8, confirming E2E-FANet's enhanced adaptability and generalization capability across varying *RW* conditions through fine-tuning.

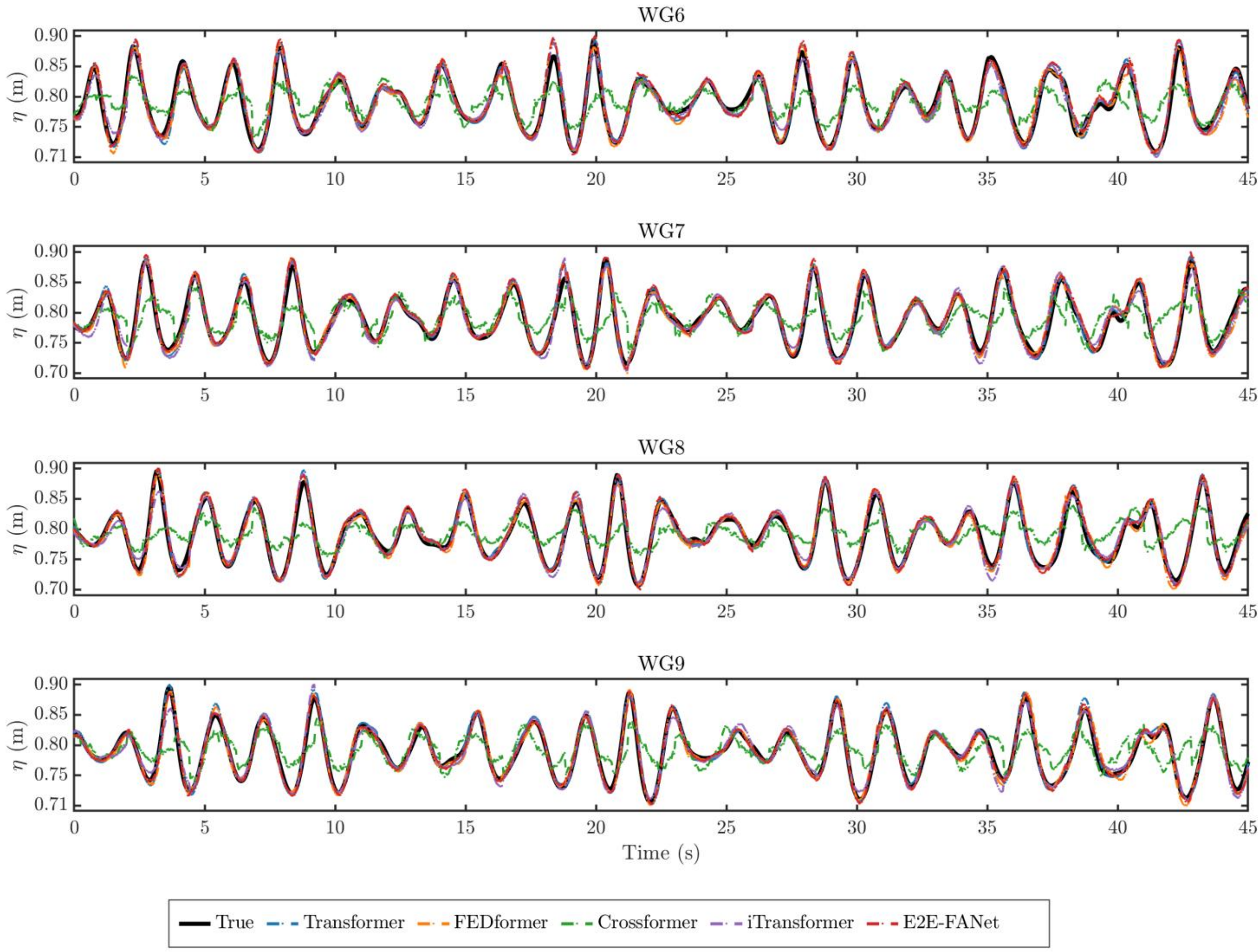

Fig. 16. Comparison of wave elevation predictions for $RW$ = 0.4 condition at downstream wave gauges.

## 5.2 Generalization results and analysis

Table 9 presents the generalization test performance of five models across varying wave conditions characterized by three parameters: $RW$, $H_s$, $T_p$. Comprehensive analysis shows that E2E-FANet consistently demonstrates superior performance across all datasets. This performance indicates that E2E-FANet demonstrates robust generalization not only across different wave conditions but also across different RW configurations. When RW changes, it alters the floating body's draft and moment of inertia, leading to changes in the floating breakwater system. This result highlights E2E-FANet's adaptability to different FB systems.

To evaluate the enhancement in generalization performance under different $RW$ conditions, Fig. 17 presents comparative bar charts of performance metrics for the three best-generalizing models, E2E-FANet, Transformer, and FEDformer, under RW=0.4 and RW=0.8 across various wave conditions. Blue bars represent $RW$ = 0.4 and red bars represent $RW$ = 0.8. Fig. 18 complements this analysis with box plots of the prediction error distributions for these models and RW conditions, providing detailed error characteristics.

The performance metrics for all three models mostly exhibit lower values at $RW$ = 0.4 compared to $RW$ = 0.8. This pattern is evident across most subplots in Fig. 17. Specifically, E2E-FANet exhibits the lowest error values under both $RW$ conditions and shows notable improvement from $RW$ = 0.8 to $RW$ = 0.4. Similarly, Transformer and FEDformer demonstrate reduced error values at $RW$ = 0.4 compared to $RW$ = 0.8. These observations indicate reduced error magnitudes at lower $RW$ across models, with E2E-FANet maintaining superior performance in both conditions. The error distribution visualization in Fig. 18 reinforces these findings. As shown in Fig. 18, blue box plots represent $RW$ = 0.4 and red box plots represent $RW$ =

0.8. The blue box plots typically display smaller interquartile ranges and shorter whiskers than their red counterparts, indicating more concentrated error distributions and thus more reliable predictions under lower *RW* conditions. Transformer and FEDformer demonstrate clear improvements in error distribution when transitioning from *RW* = 0.8 to *RW* = 0.4. The consistent trends observed in both Fig. 17 and 18 provide strong visual validation of enhanced model generalization performance under lower *RW* conditions. This is because the pre-trained model was trained on a dataset with *RW* = 0.5, where wave overtopping rarely occurs, resulting in insufficient model learning of overtopping features. While in the *RW* = 0.8 condition, wave overtopping occurs frequently, leading to larger prediction errors for this feature. Under *RW* = 0.4 condition, wave overtopping does not occur, so generalization accuracy is higher in this situation.

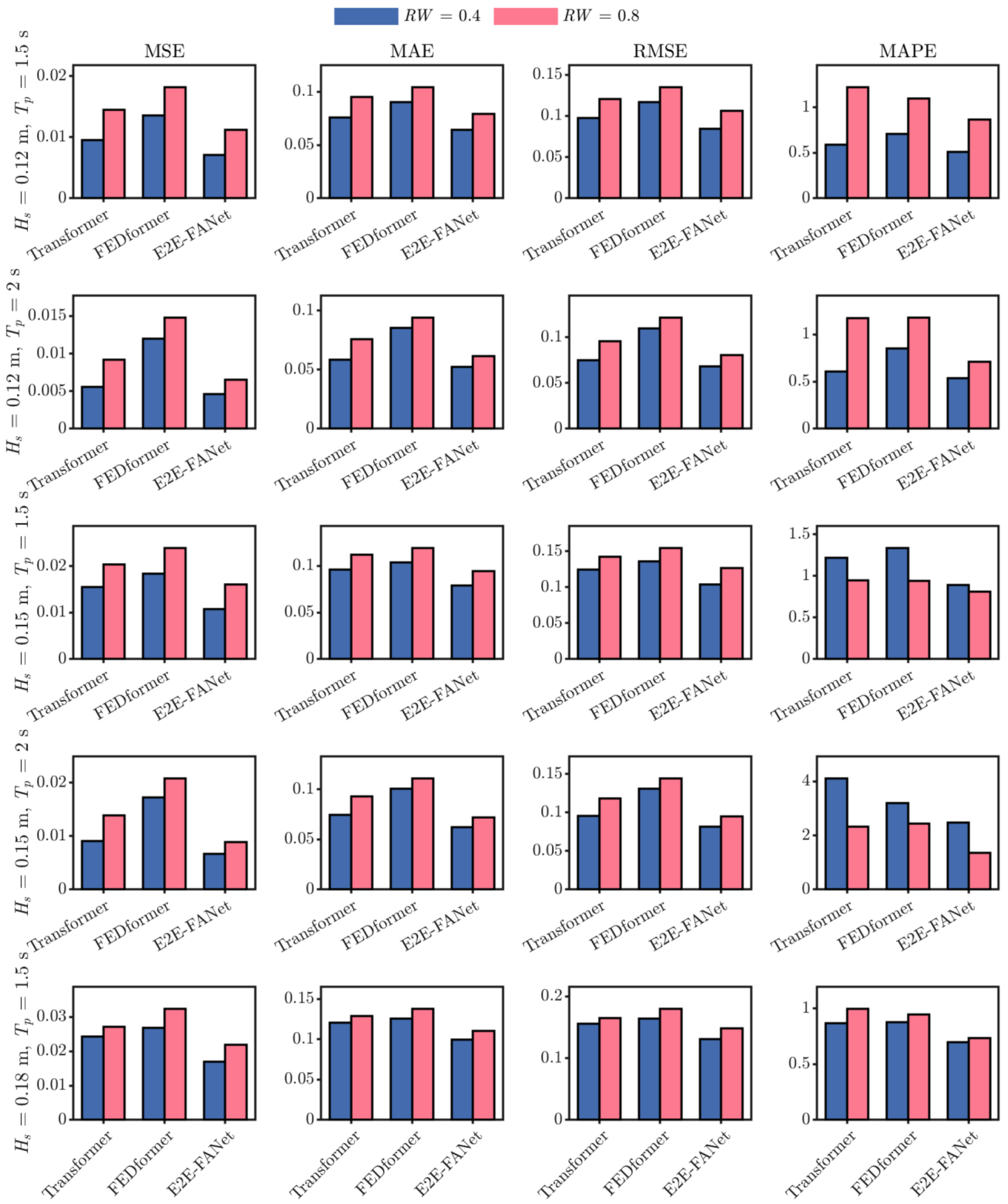

Fig. 17. Bar chart comparison of performance metrics for top generalizing models under *RW* = 0.4 and *RW* = 0.8.

Table 9 Comparative performance under different wave conditions.

| Datasets | Metric | Transformer | FEDformer | Crossformer | iTransformer | E2E-FANet |
|---|---|---|---|---|---|---|
| $RW$ = 0.8, $H_s$ = 0.12 m, $T_p$ = 1.5 s | MSE | <u>0.0145</u> | 0.0182 | 0.4557 | 0.0305 | **0.0112** |
| | MAE | <u>0.0950</u> | 0.1044 | 0.5348 | 0.1261 | **0.0795** |
| | RMSE | <u>0.1203</u> | 0.1349 | 0.6751 | 0.1747 | **0.1060** |
| | MAPE | 1.2227 | <u>1.0952</u> | 3.5742 | 1.2798 | **0.8640** |
| $RW$ = 0.8, $H_s$ = 0.12 m, $T_p$ = 2 s | MSE | <u>0.0092</u> | 0.0148 | 0.2891 | 0.0167 | **0.0065** |
| | MAE | <u>0.0756</u> | 0.0941 | 0.4175 | 0.0979 | **0.0615** |
| | RMSE | <u>0.0958</u> | 0.1215 | 0.5377 | 0.1292 | **0.0804** |
| | MAPE | 1.1710 | 1.1798 | 3.8205 | <u>0.9820</u> | **0.7128** |
| $RW$ = 0.8, $H_s$ = 0.15 m, $T_p$ = 1.5 s | MSE | <u>0.0203</u> | 0.0239 | 0.6328 | 0.0473 | **0.0160** |
| | MAE | <u>0.1121</u> | 0.1191 | 0.6352 | 0.1582 | **0.0945** |
| | RMSE | <u>0.1423</u> | 0.1545 | 0.7955 | 0.2174 | **0.1265** |
| | MAPE | 0.9499 | <u>0.9388</u> | 2.9502 | 1.0566 | **0.8089** |
| $RW$ = 0.8, $H_s$ = 0.15 m, $T_p$ = 2 s | MSE | <u>0.0139</u> | 0.0208 | 0.4171 | 0.0249 | **0.0089** |
| | MAE | <u>0.0929</u> | 0.1111 | 0.5002 | 0.1193 | **0.0719** |
| | RMSE | <u>0.1180</u> | 0.1443 | 0.6458 | 0.1577 | **0.0946** |
| | MAPE | 2.3278 | 2.4305 | 14.8757 | <u>2.3127</u> | **1.3564** |
| $RW$ = 0.8, $H_s$ = 0.18 m, $T_p$ = 1.5 s | MSE | <u>0.0271</u> | 0.0324 | 0.9670 | 0.0594 | **0.0219** |
| | MAE | <u>0.1289</u> | 0.1380 | 0.7826 | 0.1796 | **0.1103** |
| | RMSE | <u>0.1646</u> | 0.1799 | 0.9834 | 0.2437 | **0.1479** |
| | MAPE | 0.9969 | <u>0.9441</u> | 2.9188 | 1.1552 | **0.7344** |
| $RW$ = 0.4, $H_s$ = 0.12 m, $T_p$ = 1.5 s | MSE | <u>0.0095</u> | 0.0136 | 0.5081 | 0.0428 | **0.0071** |
| | MAE | <u>0.0758</u> | 0.0904 | 0.5578 | 0.1456 | **0.0641** |
| | RMSE | <u>0.0976</u> | 0.1166 | 0.7128 | 0.2069 | **0.0841** |
| | MAPE | <u>0.5903</u> | 0.7089 | 2.6783 | 0.9681 | **0.5084** |
| $RW$ = 0.4, $H_s$ = 0.12 m, $T_p$ = 2 s | MSE | <u>0.0056</u> | 0.0120 | 0.3116 | 0.0181 | **0.0046** |
| | MAE | <u>0.0584</u> | 0.0854 | 0.4382 | 0.1017 | **0.0523** |
| | RMSE | <u>0.0750</u> | 0.1097 | 0.5582 | 0.1346 | **0.0681** |
| | MAPE | <u>0.6089</u> | 0.8530 | 3.1885 | 0.8804 | **0.5339** |
| $RW$ = 0.4, $H_s$ = 0.15 m, $T_p$ = 1.5 s | MSE | <u>0.0155</u> | 0.0184 | 0.7013 | 0.0656 | **0.0108** |
| | MAE | <u>0.0957</u> | 0.1035 | 0.6604 | 0.1817 | **0.0788** |
| | RMSE | <u>0.1247</u> | 0.1358 | 0.8374 | 0.2562 | **0.1041** |
| | MAPE | <u>1.2168</u> | 1.3371 | 5.2679 | 1.8898 | **0.8912** |
| $RW$ = 0.4, $H_s$ = 0.15 m, $T_p$ = 2 s | MSE | <u>0.0091</u> | 0.0172 | 0.4564 | 0.0256 | **0.0067** |
| | MAE | <u>0.0744</u> | 0.1010 | 0.5291 | 0.1212 | **0.0625** |
| | RMSE | <u>0.0956</u> | 0.1311 | 0.6756 | 0.1600 | **0.0817** |
| | MAPE | 4.1086 | <u>3.1971</u> | 9.5926 | 3.2344 | **2.4776** |
| $RW$ = 0.4, $H_s$ = 0.18 m, $T_p$ = 1.5 s | MSE | <u>0.0243</u> | 0.0268 | 1.0304 | 0.0754 | **0.0170** |
| | MAE | <u>0.1209</u> | 0.1260 | 0.8069 | 0.1995 | **0.0996** |
| | RMSE | <u>0.1559</u> | 0.1638 | 1.0151 | 0.2747 | **0.1306** |
| | MAPE | <u>0.8679</u> | 0.8781 | 2.8270 | 1.3551 | **0.6960** |

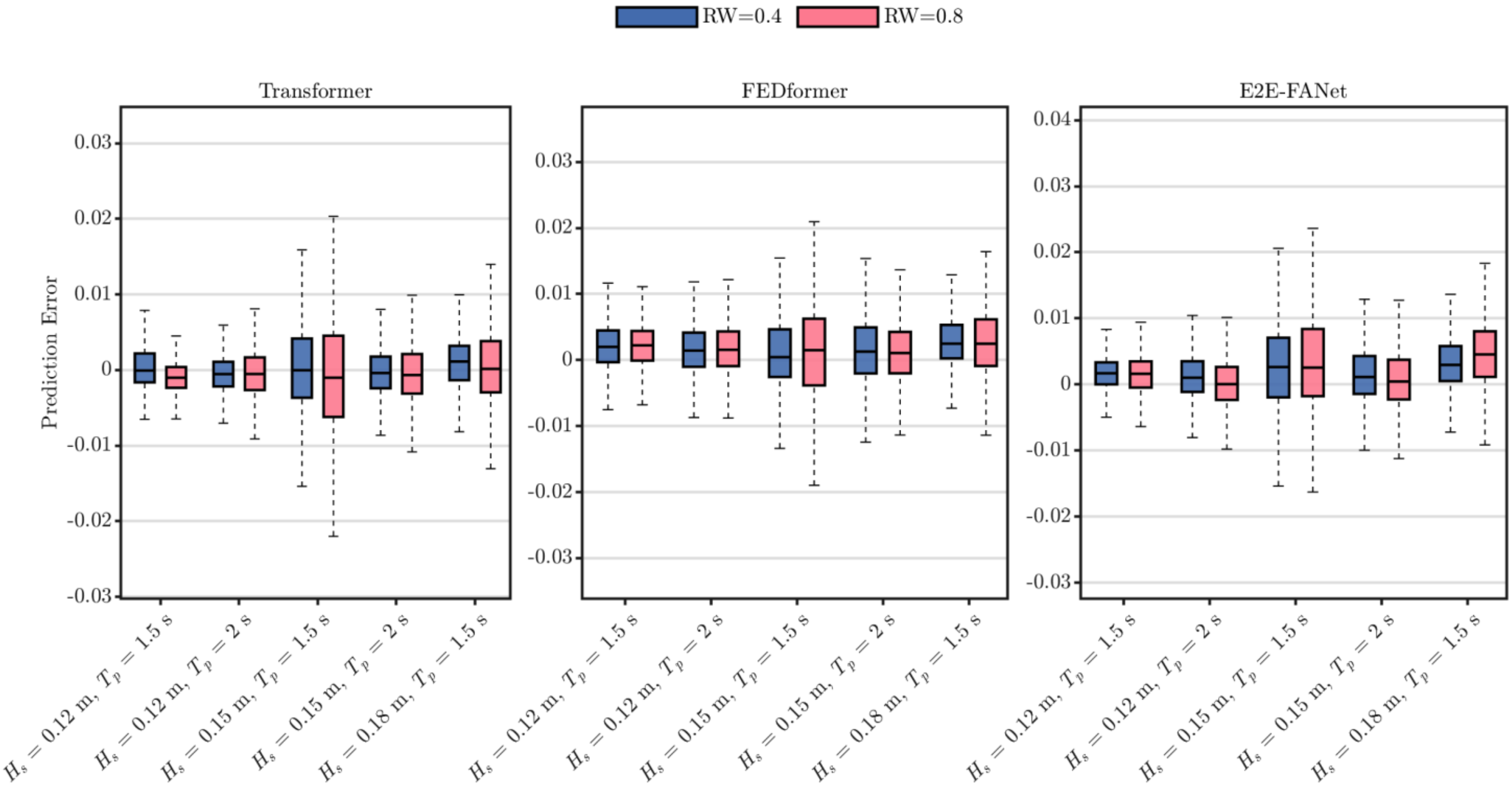

Fig. 18. Box plot comparison of prediction error distribution for top generalizing models under $RW = 0.4$ and $RW = 0.8$.

## 6. Conclusions

This study introduces E2E-FANet, a novel end-to-end deep learning framework that improves the accuracy and generalization capabilities of waves prediction behind FBs. The framework incorporates three key modules: DBFM, TA, and E2ECA. The DBFM module comprehensively extracts frequency-domain features from wave signals while preserving temporal sequence information by combining sine and cosine basis functions. The TA module captures long-term dependencies in time series data, while the E2ECA module effectively models complex interactions between endogenous and exogenous variables. Through these components, E2E-FANet efficiently captures both time-frequency characteristics and variable interactions in waves predictions behind FB, making it particularly effective for wave elevations forecasting and generalization tasks. The main conclusions are summarized as follows:

(1) E2E-FANet demonstrates superior prediction accuracy compared to baseline models across all downstream wave gauge. Both time series comparisons and cumulative absolute error analyses confirm that E2E-FANet accurately captures wave dynamics at each wave gauge position. Comprehensive experimental results demonstrate that E2E-FANet achieves superior performance across multiple evaluation metrics, significantly outperforming baseline models by reducing MSE by 19.95%, RMSE by 10.64%, and MAE by 12.88%.

(2) E2E-FANet exhibits robust generalization capabilities across datasets with diverse wave conditions. Generalization tests across multiple wave condition datasets indicate that E2E-FANet effectively predicts wave elevations behind FB under varying wave conditions. The model maintains high-precision predictive performance even for wave conditions not included in the training data, validating the robustness and generalization capacity of its architecture. The exceptional generalization capability of E2E-FANet is primarily attributed to the DBFM module. Through orthogonal basis function extraction of frequency domain features, the DBFM module effectively captures fundamental characteristics of wave spectral energy distribution, enabling the model to learn robust feature representations independent of specific wave forms.

(3) E2E-FANet demonstrates effective adaptability across diverse FB systems. When RW changes occur, the dynamic characteristics of the floating breakwater system are altered. E2E-FANet rapidly adapts to new system dynamics characteristics through fine-tuning of the pre-trained model and effectively generates accurate predictions across different wave shape datasets under new *RW* conditions. This exceptional *RW* adaptability primarily stems from the effective modeling of exogenous variable influences by the E2ECA module and its ability to swiftly adjust variable interaction relationships through fine-tuning.

However, several limitations warrant consideration. First, similar to other deep learning architectures, E2E-FANet exhibits limited interpretability, which constrains comprehensive validation of its physical plausibility. Second, its complex design, which uses multiple attention mechanisms and frequency-domain analysis. This means it may not work well for real-time processing or in situations with limited computing resources. These limitations indicate promising directions for future research. The focus should be on making the model more efficient while enhancing model interpretability, and most importantly, validating E2E-FANet on more diverse datasets, including physical model experiments and real-world oceanographic data, to further solidify its generalization capabilities and practical applicability. Additionally, future work should expand the framework applicability to more diverse and complex marine environments by incorporating sea current dynamics.

## Acknowledgements

The study was supported by the China Postdoctoral Science Foundation (Certificate Number: 2024M761844), the National Natural Science Foundation of Shandong Province (No. ZR2023QA090), the National Natural Science Foundation of China (No. 52201339).

**Appendix A**

A.1 Motion response of floating body driven by fluid

The motion of a floating breakwater is subject to Newton's second law. By assuming that the floating body is rigid, the force on the floating body can be obtained by the sum of the contributions of the surrounding fluid particles. Therefore, the force per unit mass of each floating boundary particle *k* is:

$$\mathbf{f}_k = \sum_a \mathbf{f}_{ka} \tag{A.1}$$

where, $\mathbf{f}_{ka}$ is the unit mass force of the surrounding fluid particle *a* acting on the solid particle *k*. According to Newton's third law, solid and fluid particles are subject to the same amount of force and in opposite directions:

$$m_k \mathbf{f}_{ka} = -m_a \mathbf{f}_{ak} \tag{A.2}$$

According to Newton's second law, the linear and angular velocity of a floating breakwater can be expressed as:

$$M \frac{d\mathbf{V}}{dt} = \sum_k m_k \mathbf{f}_k \tag{A.3}$$

$$I \frac{d\mathbf{\Omega}}{dt} = \sum_k m_k \left(\mathbf{r}_k - \mathbf{R}_0\right) \times \mathbf{f}_k \tag{A.4}$$

where M is the mass of the floating body. $\mathbf{V}$ and $\mathbf{\Omega}$ are linear and angular velocities, respectively. $\mathbf{R}_0$ is the position of mass center. *I* is the moment of inertia. By superimposing linear velocity and angular velocity, the velocity of the floating particle is:

$$\mathbf{u}_k = \mathbf{V} + \mathbf{\Omega} \times \left(\mathbf{r}_k - \mathbf{R}_0\right) \tag{A.5}$$

A.2 Lumped mass method

The mooring line has two ends. One is a fairlead for connecting the floating body. The other is anchored to the bottom. MoorDyn uses the lumped mass method to calculate the dynamic response of the mooring system. The mooring cable is discretized into a series of points, as shown in Fig. A. 1. The mass of the mooring cable is concentrated on the discrete points. The motion of the points can be obtained by calculating the force of the points. The forces at each concentrated mass point include net weight or buoyancy force ($\boldsymbol{W}_i$), internal axial tension and damping force ($\boldsymbol{T}_{i+1/2}$, $\boldsymbol{C}_{i+1/2}$), hydrodynamic force ($\boldsymbol{D}_{pi}$, $\boldsymbol{D}_{qi}$), and the force caused by contact with water bottom ($\boldsymbol{B}_i$).

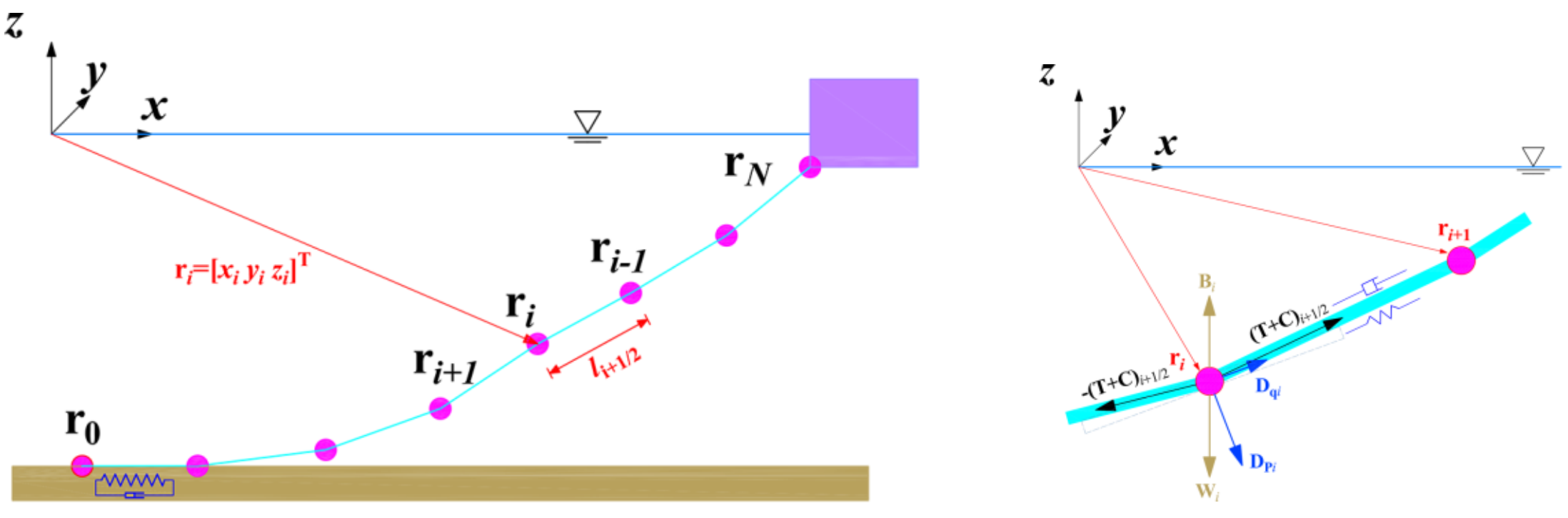

Fig. A.1 Scheme of lumped mass method

A.2.1 Net weight or buoyancy force

The net weight force or buoyancy is the difference between the gravity and buoyancy of a discrete point. When the density of water is greater than the density of the mooring line, the net buoyancy is in the positive $Z$ direction. Otherwise, it is the net weight force, and the direction is negative $Z$ direction. The net weight force of each segment $i$+1/2 is half that of two adjacent discrete points ($i$, $i$+1), which is calculated by the following formula:

$$\boldsymbol{W}_{i+1/2} = \frac{\pi}{4} d^2 l \left(\rho_m - \rho_w\right) \mathbf{g} \tag{A.6}$$

$$\boldsymbol{W}_i = \frac{1}{2}\left(\boldsymbol{W}_{i+1/2} + \boldsymbol{W}_{i-1/2}\right) \tag{A.7}$$

where $\rho_m$ and $\rho_w$ are the densities of the mooring line and water, respectively.

A.2.2 Internal axial tension and damping force

The tension of mooring segment $i$+1/2 is

$$\boldsymbol{T}_{i+1/2} = E \frac{\pi}{4} D^2 \varepsilon_{i+1/2} \left( \frac{\mathbf{r}_{i+1} - \mathbf{r}_i}{\left\| \mathbf{r}_{i+1} - \mathbf{r}_i \right\|} \right) \tag{A.8}$$

where, $E$ is the elastic modulus. $\varepsilon_{i+1/2}$ is strain. $D$ is the diameter of the mooring line. The direction of the tension is from node $i$ to $i$+1. The mooring line can only be stretched, not compressed. So, it can only have positive tension. The damping force of mooring segment $i$+1/2 is:

$$\mathbf{C}_{i+1/2} = C_{\text{int}} \frac{\pi}{4} D^2 \frac{\partial \varepsilon_{i+1/2}}{\partial t} \left( \frac{\mathbf{r}_{i+1} - \mathbf{r}_i}{\left\| \mathbf{r}_{i+1} - \mathbf{r}_i \right\|} \right) \tag{A.9}$$

where, $C_{\text{int}}$ is the internal damping coefficient. For chains with small internal damping, a small amount of structural damping will still be included to ensure the stability of the numerical calculation. In Eq. (A.9), the strain rate is calculated as in Eq. (A.10).

$$\frac{\partial \varepsilon_{i+1/2}}{\partial t} = \frac{1}{l}\frac{1}{\|\mathbf{r}_{i+1}-\mathbf{r}_i\|}\frac{\partial}{\partial t}\left[\left(x_{i+1}-x_i\right)^2+\left(z_{i+1}-z_i\right)^2\right]$$
$$= \frac{1}{l}\frac{1}{\|\mathbf{r}_{i+1}-\mathbf{r}_i\|}\left[\left(x_{i+1}-x_i\right)\left(\dot{x}_{i+1}-\dot{x}_i\right)+\left(z_{i+1}-z_i\right)\left(\dot{z}_{i+1}-\dot{z}_i\right)\right] \quad (A.10)$$

A.2.3 Hydrodynamic force

The drag force and the additional mass are obtained by Morison's equation. The relative fluid velocity and acceleration at the node are decomposed into transverse and tangential components. The tangential direction at each node is approximately the direction of the chain line through two adjacent nodes:

$$\mathbf{e}_i = \frac{\mathbf{r}_{i+1}-\mathbf{r}}{\|\mathbf{r}_{i+1}-\mathbf{r}_i\|} \quad (A.11)$$

The tangential and normal drag forces at node $i$ are:

$$\mathbf{D}_{qi} = \frac{1}{2}\rho_w C_{dt} lD \left\|\left(-\frac{\partial \mathbf{r}_i}{\partial t}\bullet\mathbf{e}_i\right)\hat{\mathbf{q}}_i\right\|\left[\left(-\frac{\partial \mathbf{r}_i}{\partial t}\bullet\mathbf{e}_i\right)\mathbf{e}_i\right] \quad (A.12)$$

$$\mathbf{D}_{pi} = \frac{1}{2}\rho_w C_{dn} lD \left\|\left(\frac{\partial \mathbf{r}_i}{\partial t}\bullet\mathbf{e}_i\right)\mathbf{e}_i - \frac{\partial \mathbf{r}_i}{\partial t}\right\|\left[\left(\frac{\partial \mathbf{r}_i}{\partial t}\bullet\mathbf{e}_i\right)\mathbf{e}_i - \frac{\partial \mathbf{r}_i}{\partial t}\right] \quad (A.13)$$

where $C_{dt}$ is the tangential drag coefficient and $C_{dn}$ is the normal drag coefficient. Tangential and normal additional mass forces are:

$$\mathbf{a}_{\mathbf{p}i}\frac{\partial^2 \mathbf{r}_i}{\partial t^2} = \rho_w C_{an}\frac{\pi}{4}d^2 l\left[\left(\frac{\partial^2 \mathbf{r}_i}{\partial t^2}\bullet\mathbf{e}\right)\mathbf{e} - \frac{\partial^2 \mathbf{r}_i}{\partial t^2}\right] \quad (A.14)$$

$$\mathbf{a}_{\mathbf{q}i}\frac{\partial^2 \mathbf{r}_i}{\partial t^2} = \rho_w C_{at}\frac{\pi}{4}d^2 l\left(-\frac{\partial^2 \mathbf{r}_i}{\partial t^2}\bullet\mathbf{e}\right)\mathbf{e} \quad (A.15)$$

where $C_{at}$ and $C_{an}$ are tangential and normal additional mass coefficients respectively. $\mathbf{a}_{\mathbf{q}i}$ and $\mathbf{a}_{\mathbf{p}i}$ are tangential and normal additional mass matrices, respectively. Therefore, the additional mass of node $i$ can be expressed as a matrix:

$$\mathbf{a}_i = \mathbf{a}_{\mathbf{p}i} + \mathbf{a}_{\mathbf{q}i} = \rho_w\frac{\pi}{4}D^2 l\left[C_{an}\left(\mathbf{I}-\mathbf{e}_i\mathbf{e}_i^{\mathrm{T}}\right)+C_{at}\mathbf{e}_i\mathbf{e}_i^{\mathrm{T}}\right] \quad (A.16)$$

A.2.4 Seabed force

The linear spring-damping method is used to simulate the forces touching the seabed:

$$\boldsymbol{B}_i = lD\left[\left(z_{bot}-z_i\right)k_b - \frac{\partial z_i}{\partial t}c_b\right] \quad (A.17)$$

where $z_{bot}$ is the position of z at the bottom of the water. $k_b$ and $c_b$ are stiff coefficients and damping coefficients respectively, representing the stiffness and viscous damping of the seabed per unit area. Seabed force is activated only when $z_i \leq z_{bot}$.

According to Newton's second law, the motion of the points on the mooring line can be expressed as:

$$\left(\mathbf{m}_i+\mathbf{a}_i\right)\ddot{\mathbf{r}}_i = \mathbf{T}_{i+1/2}-\mathbf{T}_{i-1/2}+\mathbf{C}_{i+1/2}-\mathbf{C}_{i-1/2}+\boldsymbol{W}_i+\boldsymbol{B}_i+\boldsymbol{D}_{pi}+\boldsymbol{D}_{qi} \quad (A.18)$$

The second-order Runge Kutta method with fixed time step is used to solve the problem.

## Reference

[1] Chen, J., Zhang, J., Wang, G., Zhang, Q., Guo, J., Sun, X. (2022). Numerical simulation of the wave dissipation performance of floating box-type breakwaters under long-period waves. Ocean Engineering, 266, 113091.

[2] Cebada-Relea, A. J., López, M., Claus, R., & Aenlle, M. (2023). Short-term analysis of extreme wave-induced forces on the connections of a floating breakwater. Ocean Engineering, 280, 114579.

[3] Wang, S., Xu, T. J., Wang, T. Y., Dong, G. H., & Hou, H. M. (2024). Hydrodynamic analysis of an aquaculture tank-type floating breakwater integrated with perforated baffles. Applied Ocean Research, 153, 104261.

[4] Mao, P., Chen, C., Chen, X., Zhang, Q., Bao, Y., Yang, Q. (2024). An innovative design for floating breakwater with Multi-objective genetic optimal method. Ocean Engineering, 312, 119202.

[5] Ji, C., Bian, X., Lu, L., Guo, J., Xu, S., Lv, F. (2024). 3D experimental investigation of floating breakwater with symmetrical openings and wing structures. Ocean Engineering, 313, 119624.

[6] He, Y., Han, B., Han, X., Xie, H. (2024). Diffraction wave on the single wing floating breakwater. Applied Ocean Research, 142, 103941.

[7] Han, X., Dong, S. (2023). Interaction between regular waves and floating breakwater with protruding plates: Laboratory experiments and SPH simulations. Ocean Engineering, 287, 115906.

[8] Zhang, Z., Tao, A., Wu, Q., Xie, Y. (2024). Wave to the dynamic response of the ballast floating breakwater. Ocean Engineering, 305, 117915.

[9] Cheng, Y., Xi, C., Dai, S., Ji, C., Collu, M., Li, M., Yuan, Z., Incecik, A. (2022). Wave energy extraction and hydroelastic response reduction of modular floating breakwaters as array wave energy converters integrated into a very large floating structure. Applied Energy, 306, 117953.

[10] Dai, J., Wang, C. M., Utsunomiya, T., Duan, W. (2018). Review of recent research and developments on floating breakwaters. Ocean Engineering, 158, 132-151.

[11] Saghi, H., Mikkola, T., Hirdaris, S. (2021). A machine learning method for the evaluation of hydrodynamic performance of floating breakwaters in waves. Ships and offshore Structures, 17(7), 1447-1461.

[12] Karami, H., Saghi, H. (2024). Hybrid intelligent and numerical methods to estimate the transmission coefficients of rectangular floating breakwaters. Water Supply, 24(9), 3015.

[13] Chen, J., Milne, I., Taylor, P. H., Gunawan, D., Zhao, W. (2023). Forward prediction of surface wave elevations and motions of offshore floating structures using a data-driven model. Ocean Engineering, 281, 114680.

[14] Pena-Sanchez, Y., Mérigaud, A., Ringwood, J.V., 2018. Short-term forecasting of sea surface elevation for wave energy applications: The autoregressive model revisited. IEEE Journal of Oceanic Engineering, 45 (2), 462-471.

[15] Truong, D., Ahn, K. (2012). Wave prediction based on a modified grey model MGM (1, 1) for real-time control of wave energy converters in irregular waves. Renewable Energy, 43, 242-255.

[16] Hong, W., Li, M., Geng, J., Zhang, Y. (2019). Novel chaotic bat algorithm for forecasting complex motion of floating platforms. Applied Mathematical Modelling, 72, 425-443.

[17] Shi, S., Patton, R.J., Liu, Y. (2018). Short-term wave forecasting using gaussian process for optimal control of wave energy converters. IFAC-PapersOnLine 51 (29), 44–49.

[18] Song, X., Wang, S., Hu, Z., Li, H. (2019). A hybrid Rayleigh and Weibull distribution model for the short-term motion response prediction of moored floating structures. Ocean Engineering, 182, 126-136.

[19] Wang, S., Liu, S., Xiang, C., Li, M., Yang, Z., Huang, B. (2023). Prediction of wave forces on the box-girder superstructure of the offshore bridge with the influence of floating breakwater. Journal of Marine Science and Engineering, 11(7), 1326.

[20] Duan, W., Zhang, L., Cao, S., Sun, X., Zhang, X., Huang, L. (2024). Reconstruction of significant wave height distribution from sparse buoy data by using deep learning. Coastal Engineering, 194, 104616.

[21] Wang, J., Jin, X., He, Z., Wang, Y., Liu, X., Chai, J., Guo, R. (2024). Research on high precision online prediction of motion responses of a floating platform based on multi-mode fusion. Applied Ocean Research, 151, 104150.

[22] Yuan, L., Chen, Y., Zan, Y., Zhong, S., Jiang, M., Sun, Y. (2023). A novel hybrid approach to mooring tension prediction for semi-submersible offshore platforms. Ocean Engineering, 287, Part1, 115776.

[23] Yuan, L., Chen, Y., Li, Z. (2024). Real-time prediction of mooring tension for semi-submersible platforms. Applied Ocean

Research, 146, 103967.
[24] Xu, G., Zhang, S., Shi, W. (2023). Instantaneous prediction of irregular ocean surface wave based on deep learning. Ocean Engineering, 267, 113218.
[25] Yao, Y., Han, L., & Wang, J. (2018). Lstm-pso: Long short-term memory ship motion prediction based on particle swarm optimization. In 2018 IEEE CSAA Guidance, Navigation and Control Conference (CGNCC) (pp. 1-5). IEEE.
[26] Shi, W., Hu, L., Lin, Z., Zhang, L., Wu, J., Chai, W. (2023). Short-term motion prediction of floating offshore wind turbine based on muti-input LSTM neural network. Ocean Engineering, 280, 114558.
[27] Payenda, M. A., Wang, S., Jiang, Z., Prinz, A. (2024). Prediction of mooring dynamics for a semi-submersible floating wind turbine with recurrent neural network models. Ocean Engineering, 313, 119490.
[28] Li, D., Jiang, M., Li, M., Hong, W., Xu, R. (2023). A floating offshore platform motion forecasting approach based on EEMD hybrid ConvLSTM and chaotic quantum ALO. Applied Soft Computing, 144, 110487.
[29] He, G., Xue, J., Zhao, C., Cui, T., Liu, C. (2024). Deep learning based short-term motion prediction of floating wind turbine under shutdown condition. Applied Ocean Research, 151, 104147.
[30] Ye, Y., Wang, L., Wang, Y., Qin, L. (2022). An EMD-LSTM-SVR model for the short-term roll and sway predictions of semi-submersible. Ocean Engineering, 256, 111460.
[31] Neshat, M., Nezhad, M.M., Sergiienko, N.Y., Mirjalili, S., Piras, G., Garcia, D.A., 2022. Wave power forecasting using an effective decomposition-based convolutional Bidirectional model with equilibrium Nelder-Mead optimiser. Energy. 256, 124623.
[32] Zhao, L., Li, Z., Qu, L., Zhang, J., Teng, B. (2023). A hybrid VMD-LSTM/GRU model to predict non-stationary and irregular waves on the east coast of China. Ocean Engineering, 276, 114136.
[33] Ding, T., Wu, D., Shen, L., Liu, Q., Zhang, X., & Li, Y. (2024). Prediction of significant wave height using a VMD-LSTM-rolling model in the South Sea of China. Frontiers in Marine Science, 11, 1382248.
[34] Lin, Z., Li, M., Zheng, Z., Cheng, Y., Yuan, C. (2020). Self-attention convlstm for spatiotemporal prediction. In: Proceedings of the AAAI Conference on Artificial Intelligence, 34(07), 11531-11538.
[35] Parisotto, E., Song, F., Rae, J., Pascanu, R., Gulcehre, C., Jayakumar, S., Jaderberg, M., Kaufman, R. L., Clark, A., Noury, S., Botvinick, M. M., Heess, N., Hadsell, R. (2020). Stabilizing transformers for reinforcement learning, in: International Conference on Machine Learning, PMLR, pp. 7487–7498.
[36] Song, D., Dai, S., Li, W., Ren, T., Wei, Z., Liu, A. (2024). STVformer: Aspatial-temporal-variable transformer with auxiliary Knowledge for sea surface temperature prediction. Applied Ocean Research, 153, 104218.
[37] Wu, H., Xu, J., Wang, J., & Long, M. (2021). Autoformer: Decomposition transformers with auto-correlation for long-term series forecasting. Advances in neural information processing systems, 34, 22419-22430.
[38] Vaswani, A., Shazeer, N., Parmar, N., Uszkoreit, J., Jones, L., Gomez, A. N., ... & Polosukhin, I. (2017). Attention is all you need. Advances in neural information processing systems, 30.
[39] Yang, R., Cao, L., & YANG, J. (2024). Rethinking fourier transform from a basis functions perspective for long-term time series forecasting. Advances in Neural Information Processing Systems, 37, 8515-8540.
[40] Liu, Y., Hu, T., Zhang, H., Wu, H., Wang, S., Ma, L., & Long, M. (2023). itransformer: Inverted transformers are effective for time series forecasting. arXiv preprint arXiv:2310.06625.
[41] Li, J., Selvaraju, R., Gotmare, A., Joty, S., Xiong, C., & Hoi, S. C. H. (2021). Align before fuse: Vision and language representation learning with momentum distillation. Advances in neural information processing systems, 34, 9694-9705.
[42] Hochreiter, S. (1997). Long Short-term Memory. Neural Computation MIT-Press.
[43] Bai, S., Kolter, J. Z., & Koltun, V. (2018). An empirical evaluation of generic convolutional and recurrent networks for sequence modeling. arXiv preprint arXiv:1803.01271.
[44] Zhou, T., Ma, Z., Wen, Q., Sun, L., Yao, T., Yin, W., & Jin, R. (2022). Film: Frequency improved legendre memory model for long-term time series forecasting. Advances in neural information processing systems, 35, 12677-12690.
[45] Wu, H., Yang, C., & Hu, H. (2023). Frequency Enhanced Time Series Forecasting. ACM Transactions on Knowledge Discovery from Data (TKDD), 17(5), 1-24.
[46] Zhang, Z., Xu, Y., Xiao, W., Zhang, L., Liu, H., Wang, W., ... & Zhou, J. (2023). Crossformer: Transformer utilizing cross-dimension dependency for multivariate time series forecasting. Proceedings of the AAAI Conference on Artificial Intelligence, 37(2), 1368-1376.

[47] Liu, K., Jin, H., Wang, Y., Xie, X., Zhang, S., Liu, H., ... & Cheng, H. (2023). iTransformer: Inverted transformer for time series forecasting. International Conference on Learning Representations.
[48] Zhou, T., Wu, S., Hou, X., & Zhang, Y. (2022). FEDformer: Frequency enhanced transformer for long-term time series forecasting. International conference on machine learning, 26723-26741.